\documentclass[Afour,sagev,times]{sagej}

\usepackage{moreverb,url}

\usepackage[colorlinks,bookmarksopen,bookmarksnumbered,allcolors=black]{hyperref}

\newcommand\BibTeX{{\rmfamily B\kern-.05em \textsc{i\kern-.025em b}\kern-.08em
T\kern-.1667em\lower.7ex\hbox{E}\kern-.125emX}}

\usepackage{amssymb}
\usepackage{amsmath}
\usepackage{booktabs}
\usepackage[Algorithm,ruled]{algorithm}
\usepackage{algorithmic}
\usepackage[acronym]{glossaries} 
\usepackage{adjustbox}
\usepackage{caption}
\usepackage{subcaption}
\usepackage{makecell}
\usepackage{longtable} 
\usepackage{adjustbox}
\usepackage[table,xcdraw]{xcolor}
\usepackage{comment}
\usepackage{multirow}
\usepackage[justification=centering]{caption}
\usepackage{array}
\usepackage{ulem}
\usepackage{xcolor}
\usepackage{orcidlink}

\glsdisablehyper
\definecolor{verde}{rgb}{0.0, 0.6, 0.0}
\definecolor{rojo}{rgb}{0.96, 0.76, 0.76}

\def\hb{\hbox to 11.5 cm{}}

\newcommand{\silence}[1]{}

\newacronym{SERA}{SERA}{Single European Railway Area}
\newacronym{IM}{IM}{Infrastructure Manager}
\newacronym[firstplural = Railway Undertakings (RUs)]{RU}{RU}{Railway Undertaking}
\newacronym{GA}{GA}{Genetic Algorithm}
\newacronym{PSO}{PSO}{Particle Swarm Optimization}
\newacronym{ACO}{ACO}{Ant Colony Optimization}
\newacronym{ACOR}{ACOR}{Ant Colony Optimization Continuous}
\newacronym{SA}{SA}{Simulated Annealing}
\newacronym{DE}{DE}{Differential Evolution}
\newacronym{CMA-ES}{CMA-ES}{Covariance Matriz Adaptation}
\newacronym{ABC}{ABC}{Artificial Bee Colony}
\newacronym{GWO}{GWO}{Grey Wolf Optimize}
\newacronym{TS}{TS}{Tabu Search}
\newacronym{HC}{HC}{Hill Climbing}
\newacronym{WOA}{WOA}{Whale Optimization Algorithm}
\newacronym{GWO-WOA}{GWO-WOA}{Hybrid Grey Wolf - Whale Optimization Algorithm}
\newacronym{URM}{URM}{Uniform Rectilinear Motion}
\newacronym{TTP}{TTP}{Train Timetabling Problem}
\newacronym{SCIP}{SCIP}{Solving Constraint Integer Programs}

\begin{document}

\title{An approach to the timetabling problem in deregulated railway markets based on metaheuristic algorithms}

\author{
David Muñoz-Valero\affilnum{1} \orcidlink{0000-0002-2509-9911},
Juan Moreno-Garcia\affilnum{1} \orcidlink{0000-0003-2430-145X},
Julio Alberto López-Gómez\affilnum{2} \orcidlink{0000-0001-6291-6637},
Enrique Adrian Villarrubia-Martin\affilnum{2} \orcidlink{0009-0002-8006-5711}
and 
Luis Rodriguez-Benitez\affilnum{2} \orcidlink{0000-0002-7665-943X}%
}

\affiliation{\affilnum{1}Department of Technologies and Information Systems, Universidad de Castilla-La Mancha, Avenida Carlos III, s/n, Toledo, 45071, Spain \\
\affilnum{2}Department of Technologies and Information Systems, Universidad de Castilla-La Mancha, Paseo de la Universidad 4, Ciudad Real, 13071, Spain}

\corrauth{Luis Rodriguez-Benitez, \href{mailto:luis.rodriguez@uclm.es}{luis.rodriguez@uclm.es}}

\begin{abstract}
The train timetabling problem in liberalized railway markets represents a challenge to the coordination between infrastructure managers and railway undertakings. Efficient scheduling is critical to maximizing infrastructure capacity and utilization while adhering as closely as possible to the requests of railway undertakings. These objectives ultimately contribute to maximizing the infrastructure manager's revenues. This paper sets out a modular simulation framework to reproduce the dynamics of deregulated railway systems. Ten metaheuristic algorithms using the \textit{MEALPY} Python library are then evaluated in order to optimize train schedules in the liberalized Spanish railway market. In addition, an analysis of the scalability of the problem has been carried out by comparing the results with those obtained with a classical mathematical model such as SCIP in Pyomo. The results show that the Genetic Algorithm outperforms others in revenue optimization, convergence speed, and schedule adherence. Alternatives, such as Particle Swarm Optimization and Ant Colony Optimization Continuous, show slower convergence and higher variability. The results emphasize the trade-off between scheduling more trains and adhering to requested times, providing insights into solving complex scheduling problems in deregulated railway systems.
\end{abstract}


\maketitle

\section{Introduction}
\label{sec:introduction}
In recent decades, the railway system in European Union countries has undergone major evolution \cite{Ait2021}, and the main challenge of all the changes the sector is facing is the establishment of a \gls{SERA} \cite{COM2011}. Achieving this goal will have two important consequences: on the one hand, it will open up the passenger rail market to new operating \glspl{RU}. This will lead both to an increase in the supply available to users and to more competitive prices \cite{Caramello2017}. On the other hand, a second consequence of the creation of \gls{SERA} will allow railway operators to provide passenger services in every member state \cite{Caramello2017}. Currently, projects such as the Timetabling and Capacity Redesign for Smart Capacity Management (TTR), seek to improve the flexibility, efficiency and cost-effectiveness of the market to meet today's needs \footnote{Forum Train Europe FTE, \textit{Introduction to TTR}: \url{https://www.forumtraineurope.eu/services/ttr}} . For example, in Switzerland, capacity planning under this program is based on the concepts of network usage (NNK) and network usage plan (NNP), aligning the capacity assignment strategies with the existing national tools available and the neighboring countries. The gradual transition towards a liberalized European rail market affects the market structure in the member states \cite{Ait2021}. Currently, there is a vertical separation in railway markets \cite{Bouf05}. This structure defines three clearly differentiated players or roles in the market: the \gls{IM}, the \glspl{RU} and the demand \cite{Besinovic2024}. The \gls{IM} is the entity that, through a contract with the owner of the railway infrastructure, obtains the rights to manage it. Some of its functions include the capacity allocated to the different \glspl{RU}, track access charging, or infrastructure maintenance, among others \cite{Pena2015}. The \glspl{RU} are the companies offering rail services. In a liberalized railway market, there is more than one \gls{RU} and they compete with each other in two ways: firstly, they compete for the infrastructure managed by the \gls{IM} and, secondly, they compete on price to attract passengers, the third main actor in the market, corresponding to the demand \cite{Link04}. This contrasts with the classic monopolistic framework, where a single \gls{RU} both plans and operates services under unified control, greatly simplifying the timetabling process. Whereas traditional timetabling assumes centralized decision-making, liberalized markets require coordination, strategic bidding for slots and dynamic pricing considerations. In this current context, it is crucial to develop advanced tools to solve the complex problems that arise in this dynamic and competitive environment. These include the allocation of time slots to \glspl{RU} \cite{Besinovic2024}, the establishment of timetables, the setting of prices \cite{Smoliner2018}, among others. This proposal specifically focuses on solving the timetabling problem in a liberalized railway market. 


\subsection{Contributions}
\label{subsec:contributions}

This paper focuses on the railway timetabling problem in liberalized railway markets. Specifically, the Spanish railway market has been used as a test case for the computational experiments presented in this study. The main contributions of this paper are the following:

\begin{itemize}
  \item \textbf{A novel \gls{RU} competitive timetabling model.} This study introduces a dynamic mathematical formulation that models the interactions between \gls{IM} and \glspl{RU} in a realistic and adaptable way. Its main innovation lies in the development of an analytical tool capable of simulating multiple scenarios and policy options, paving the way for automated, data-driven decision making in the efficient allocation of resources within highly competitive and dynamic railway markets.

  \item \textbf{A heuristic for scheduling conflicting services: } The paper introduces a deterministic greedy algorithm that converts the continuous departure‐time vector produced by the metaheuristics into a discrete, conflict‐free service schedule. The heuristic first identifies and schedules all services without conflicts, then iteratively selects the highest‐revenue service among those in conflict, adds it to the schedule, and removes its conflicting services from further consideration. By doing so, it reduces the combinatorial complexity, delivering consistent, high‐quality schedules even for large service sets.
  

  \item \textbf{Customized bio-inspired optimization framework.} This study advances the state of the art by leveraging the open-source \textit{MEALPY} library to apply metaheuristic and bio-inspired algorithms to the complex timetabling problem in a liberalized railway market—an area where such techniques have been under-explored. Unlike prior work, which often lacks transparency or adaptability, this research systematically adapts and integrates these algorithms—proven effective in other scheduling domains—into a replicable and consistent frameworkwhich is publicly available in the following repository\footnote{\url{https://github.com/DavidMunozValero/GSA_M}} to ensure that the results are reproducible. By building on \textit{MEALPY} \cite{Vanthieu2023,Vanthieu2023b}, the study not only demonstrates the applicability of these methods to railway operations, but also introduces a robust, open computational environment that facilitates experimentation and benchmarking in complex optimization scenarios.
  

  \item \textbf{Metaheuristic comparison under uniform conditions.} This study makes a novel contribution by systematically benchmarking a diverse set of metaheuristic algorithms—from classical to more recent approaches—under identical experimental conditions. Unlike previous research, which often evaluates algorithms in isolation or under varying setups, this work provides a unified and rigorous comparison framework. It analyzes not only solution quality and runtime, but also convergence behavior and sensitivity to hyperparameters, offering a comprehensive view of algorithmic performance. Furthermore, a comparison with a classical mathematical approach has been carried out by using Pyomo \cite{hart2011pyomo} and the \gls{SCIP} solver \cite{scip}. The findings reveal meaningful trade-offs between quality, efficiency, and robustness, offering practical guidance for algorithm selection in railway timetabling. The direct, side-by-side comparison of these algorithms in this specific context represents a significant advance over the existing literature.

\end{itemize}

\subsection{Structure of the paper}
The paper is structured as follows: Section \ref{sec:Background} reviews metaheuristic algorithms in scheduling, and recent contributions in transport and railway planning. Section \ref{sec:FormulationAndSimulation} formulates the mathematical model for the timetabling problem. Section \ref{sec:results} analyzes the results of the algorithms applied. Finally, Section \ref{sec:Conclusions} presents conclusions and future lines of research.

\section{Background}
\label{sec:Background}

A metaheuristic algorithm is a high-level algorithmic framework that provides a set of strategies for developing heuristic optimization algorithms independently of the problem to be solved \cite{Sorensen13}. To develop this type of algorithm, the scientific community has been inspired by a multitude of sources of knowledge, such as the theory of evolution, biological systems, the behavior of swarms of species, etc \cite{delSer19}. Thanks to this, a large number of metaheuristic algorithms have appeared which, depending on their source of inspiration, have been classified as evolutionary algorithms, physics-based algorithms, bio-inspired, etc \cite{Talbi02, Rajwar23}. These algorithms can return good solutions to complex problems in a shorter time than exact algorithms, and are thus very useful for solving optimization problems that require good solutions in a short time.

The following sections contain a literature review regarding the railway timetabling approaches and the applied metaheuristics algorithms.

\subsection{Railway timetabling approaches}

Metaheuristic algorithms have been successfully applied to solving scheduling and planning problems in many fields in the literature \cite{Zarandi20}. These applications range from production systems \cite{Neufeld23} to process planning \cite{Li15} and electrical power systems \cite{Morquecho24}. This remarkable success has led the transport community to focus on these methods as an alternative to the exact mathematical methods commonly used in this field of application. Thus, this section explores the application and approaches used in the literature to the solution of the railway timetabling problem.

Railway timetabling is a hot topic in the transport and operations research communities. Many contributions and breakthroughs have been made in this field in the last fifteen years. Traditionally, optimal railway timetables were calculated to minimize passenger waiting times. However, this waiting-time-focused approach neglected to deal with train delays, resulting in timetables that were not as efficient as they should have been. In \cite{Liebchen2010}, an integer-linear programming model was proposed to compute delay resistant railway timetables to fill this research gap. Derived from train delays, canceled connections are another factor affecting the efficiency of timetables. Based on the previous work, \cite{Corman2012} developed a multi-objective optimization approach to compute optimal timetables minimizing train delays and missed connections. This approach allows it to be considered which connection to keep or drop in order to reach a compromise solution. During these years, many optimization models emerged for timetable optimization and real-time re-planning of railway services. A wide review of all these approaches can be found in \cite{Cacchiani2014}.

Later, dynamic demand considerations were introduced in the proposed formulations to solve the railway timetabling problem. For example, \cite{Barrena2014} proposed two mathematical programming formulations considering a dynamic demand pattern in a single-line rail rapid transit system. Furthermore, the authors proposed a heuristic tailored algorithm to solve large instances of this problem, improving the results obtained by other solvers like CPLEX. The same authors, in \cite{Barrena2014_2}, developed three mathematical programming formulations for the train timetabling problem with dynamic demand, and the problem was solved by means of a branch-and-cut approach.

Considering demand in a dynamic way to solve the timetabling problem led to the formulation of the passenger-centric timetabling problem \cite{Robenek2016}. In contrast to traditional approaches where the aim was to minimize waiting times and delays, the passenger-centric timetabling problem aims to maximize customer satisfaction in timetable design. However, the methods for its resolution were still mixed integer linear programming models. Then, more optimization models based on this approach were proposed. A good example is the work of \cite{Binder2017}, where the authors proposed a multi-objective optimization model in order to maximize passenger satisfaction, and minimize the operational costs and the deviation from an undisrupted timetable. Other recent work is \cite{Cacchiani2020}, where a mixed-integer linear programming model was introduced to solve the timetabling and stop planning problem considering an uncertain demand.

With respect to liberalized railway markets, the work of \cite{Wong10} was a first approach to modeling intelligent negotiation between the \gls{IM} and the \glspl{RU} in a liberalized railway market. A multi-agent reinforcement learning framework based on a Q-Learning algorithm was proposed for this purpose. The simulation results provided by these authors showed that their approach enhances the intelligent behavior of the actors involved in pursuing their objectives. Then, in \cite{Ho12}, the \gls{PSO} algorithm was used to solve the timetabling problem in a liberalized railway market. This paper, unlike the previous one, focused more on the solution of the timetabling problem than on the bargaining between market players. The simulation experiments carried out by the authors allowed them to determine that \gls{PSO} was capable of generating feasible timetables that allowed the \gls{IM} to maximize both track access charges and capacity use. Game theory is another approach used to solve the railway planning problem. In \cite{Shakibaei19}, game theory allowed not only the problems of interaction/negotiation between different agents to be solved, but also the behavior of each of these agents, who want to maximize their own objectives or interests, to be identified and understood. Taking this paper as a starting point, these same authors developed, in \cite{Shakibaei21}, a multi-agent system based on a cooperative game-theoretic non-transferable utility approach, in order to solve the timetabling problem. The results obtained by the authors in a simulated environment of the Istanbul-Ankara high-speed rail network provided a Nash solution with profitable services and an acceptable level of passenger satisfaction. The problem of timetabling is not a simple one, and even less so in a liberalized environment. In \cite{Gestrelius20}, a total of seven quality aspects that a timetable should have are identified and a study is carried out in which eight Swedish practitioners were interviewed to find out the current state of practice in that country.

Finally, there are different recent studies for train scheduling in complex railway stations, tackling aspects such as track assignment, coordination, and conflict minimization. For example, \cite{Zhong24} proposed a parallel optimization method for the simultaneous scheduling of trains and shunting operations at complex high-speed railway stations. In addition, \cite{Liu23} developed a control strategy for the stable formation of high-speed virtually coupled trains under disturbances and delays, ensuring synchronized movements and reduced headway conflicts— an approach that can further enhance overall scheduling robustness and line capacity use.

In summary, the train timetabling problem has traditionally been tackled using mathematical programming methods like integer and mixed-integer linear programming. While some studies have applied metaheuristic algorithms tailored to the problem, there is still a gap in understanding their usefulness for this problem. Metaheuristics offer advantages such as being derivative-free, easy to implement, and capable of providing quick, approximate solutions— especially valuable in complex or time-sensitive scenarios. 
This paper attempts to address this research gap by designing a framework for using any metaheuristic algorithm (not necessarily designed purposely for this problem). Furthermore, this work aims to address that gap by systematically evaluating ten metaheuristic algorithms to establish a baseline for their potential and suitability in solving the train timetabling problem. Then, the performance of metaheuristic algorithms is compared with exact mathematical methods in order to provide a benchmark for the timetabling problem and understand the main advantages of the proposed approach.

\subsection{Metaheuristic algorithms of study}

This section details the metaheuristic algorithms applied to solving the train timetabling problem in a liberalized railway system. For each, their implementation in the Python \textit{MEALPY} library has been taken as a reference, since it is known for its comprehensiveness and citation impact. A representative set of ten metaheuristic algorithms was selected based on three criteria: (1) Historical coverage, including both classical and recent algorithms, from Simulated Annealing in 1985 to a Hybrid Grey Wolf–Whale Optimization Algorithm in 2022; (2) Diversity of types, covering single-individual algorithms (such as SA), those population-based, including evolutionary (GA, DE, CMA-ES), bio-inspired (ACOR, PSO, ABC, GWO, WOA) and hybrid algorithms, like the GWO-WOA; (3) Performance, with all algorithms recognized for their effectiveness and literature relevance, and many of them (GA, PSO, DE and CMA-ES) provide the basis of algorithms that have frequently won major international optimization competitions, such as IEEE CEC \cite{Qiao24} and GECCO\footnote{\url{https://www.gecad.isep.ipp.pt/ERM-competitions/2024-2/}}.

In summary, Table \ref{tab:algorithms_combined} shows each of the algorithms used, with a brief description, and their references, including the applications of each algorithm in transport problems (if known).

\begin{table*}[ht]
\centering
\scriptsize
\caption{Metaheuristic algorithms, their descriptions, general references, and transport-specific applications.}
\begin{tabular}{|>{\centering\arraybackslash}p{2.5cm}|p{7.5cm}|>{\centering\arraybackslash}p{2cm}|>{\centering\arraybackslash}p{2cm}|}
\hline
\textbf{Algorithm} & \textbf{Description} & \textbf{References} & \textbf{Transport Applications} \\ \hline \hline
Genetic Algorithm (GA) & Inspired by natural selection and genetics. Evolves a population through selection, crossover, and mutation. Widely applied in railway scheduling, rolling stock circulation, and timetabling. & \cite{Holland75} & \cite{Canca18, Pan20, Han21} \\ \hline
Simulated Annealing (SA) & Mimics the annealing process in metallurgy. Explores the solution space probabilistically, accepting worse solutions to escape local optima. Used in timetable design, maintenance scheduling, and rolling stock problems. & \cite{Cerny85, Kirkpatrick85} & \cite{Yue17, Robenek18, Lin19} \\ \hline
Ant Colony Optimization Continuous (ACOR) & Extends the ACO paradigm to continuous optimization. Simulates the pheromone-laying behavior of ants. Applied in train routing, rolling stock planning, and timetabling. & \cite{Dorigo92} & \cite{Tsuji12, Sama16, Coviello23} \\ \hline
Particle Swarm Optimization (PSO) & Inspired by the social behavior of birds and fish. Optimizes by iteratively updating candidates based on velocity and position. Commonly used in timetable optimization. & \cite{Eberhart95} & \cite{Ho12, Jamili12, Wu16} \\ \hline
Differential Evolution (DE) & A population-based algorithm combining differences between individuals to evolve solutions. Recently applied to train delay scheduling and timetable optimization. & \cite{Storn95, Storn97} & \cite{Zhong13, Song23} \\ \hline
Artificial Bee Colony (ABC) & Simulates the foraging behavior of honey bees. Balances exploration and exploitation through employed, onlooker, and scout bees. Recently applied in timetable design. & \cite{Karaboga05, Karaboga07} & \cite{Chen22, Wang24} \\ \hline
Covariance Matrix Adaptation Evolution Strategy (CMA-ES) & Adapts the covariance matrix to guide sampling. Limited transport applications, but used in railway real-time scheduling. & \cite{Auger11} & \cite{Prasad20} \\ \hline
Grey Wolf Optimizer (GWO) & Mimics the hierarchy and hunting behavior of gray wolves. Uses alpha, beta, and delta wolves to lead the search process. Applied in railway scheduling. & \cite{Mirjalili14} & \cite{Yu23, Zhang24} \\ \hline
Whale Optimization Algorithm (WOA) & Inspired by the bubble-net feeding behavior of whales. Alternates between encircling prey and random search. Limited transport applications, included due to its recent popularity. & \cite{Mirjalili16} & - \\ \hline
Hybrid Grey Wolf - Whale Optimization Algorithm (GWO-WOA) & Combines strengths of GWO and WOA. Leverages GWO leadership and WOA encircling to enhance performance. Selected to compare hybrid and standalone algorithm efficiencies. & \cite{Obadina22} & - \\ \hline
\end{tabular}
\label{tab:algorithms_combined}
\end{table*}

\section{Formulation and simulation of the proposal}
\label{sec:FormulationAndSimulation}

The scheduling of train timetables is a complex problem to deal with in a liberalized railway market. The model of railway path request and assignment phases is part of a harmonized European approach promoted by RailNetEurope (RNE) and supported by the European Union Agency for Railways (ERA) \footnote{\url{https://www.era.europa.eu/domains/technical-specifications-interoperability/operation-and-traffic-management-tsi_en}}. This model is implemented through the Path Coordination System (PCS), an international platform that enables coordination of path requests among infrastructure managers, railway undertakings, and freight corridors, ensuring consistency between requests and capacity offers. Moreover, the PCS system complies with the technical requirements established in the TAF-TSI regulation (Technical Specifications for Interoperability for Telematics Applications), which governs the interoperability of capacity and timetable management systems across the European Union. This approach is used by countries such as Spain, Switzerland, Germany, and France, among others, particularly in the context of international traffic and annual timetable planning.

\begin{figure}[ht]
    \centering
    \includegraphics[width=0.40\textwidth]{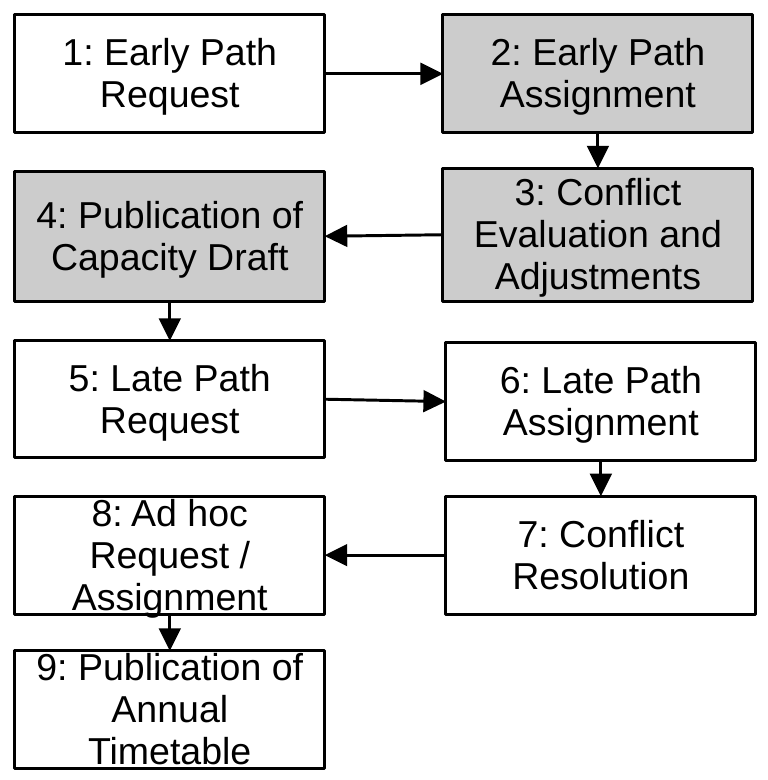}
    \caption{Relationship between the RUs and IM.}
    \label{fig:RUs_IM_relation}
\end{figure}

Figure \ref{fig:RUs_IM_relation} shows how this process plays out, using a flowchart that provides a structured representation of the key phases in the railway path request and assignment process, from capacity planning to the publication of the annual timetable. This process begins approximately eleven months before the timetable is published, during the Capacity Planning phase, where early path requests (pre-request) are submitted by the RUs and tentative assignments (pre-assignment) are made by \gls{IM} (Phases 1 and 2). Conflicts between requests are then evaluated, and available capacity is adjusted (Phase 3), culminating in the publication of a draft capacity plan (Phase 4). Subsequently, during the Timetable Generation phase, formal path requests (late request) are processed along with their corresponding definitive assignments (late assignment), phases 5 and 6 in the figure. Remaining conflicts are resolved (Phase 7), and ad hoc requests—those submitted outside standard deadlines—are handled (Phase 8). Finally, the definitive annual timetable is published, integrating all confirmed assignments (Phase 9). This approach enables a progressive and flexible management of railway capacity, adapting to the evolving needs of railway undertakings over time.


In this paper, the research carried out can be useful in phases 2 to 4, helping to optimize the process and resolve the conflicts generated. In this case, it focuses on the maximization of the \gls{IM} benefit and the payment of penalties to the \glspl{RU} affected by the changes. To solve the problem, this paper sets out an optimization problem which will be solved by means of metaheuristic algorithms. Given the difficulty of obtaining real data on this process, simulation is used here. Specifically, a modular system has been designed that replicates the previous process and generates the different elements necessary to carry out the tests. First, Section \ref{sec:optimization_problem} sets out the mathematical formulation of this optimization problem. Then, Section \ref{sec:modular_system_designed} shows the structure of the modular system designed, and finally, Section \ref{sec:general_scheme} describes the general outline of the optimization process.  

\subsection{Optimization problem}
\label{sec:optimization_problem}

With all these elements, the timetabling problem in a liberalized railway market is defined as an optimization problem. The objective function of the \gls{IM} is the maximization of the profit obtained in the planning of the services proposed by the \glspl{RU} according to Equation \ref{ecu:problem_formulation}. If the operating times of a service $ i $ do not achieve the minimum operating times based on the requested times, or exceed the operating times by a specific margin, service $ i $ would automatically not be scheduled (i.e. no revenue will be obtained with service $ i $). 

\begin{equation} \label{ecu:problem_formulation}
{\scriptstyle\max \left[ \sum_{i=1}^{|SS|} B_i \cdot \left( ca_i \cdot \left( 1 - \alpha_{i} \cdot P_{DT_{i}} - \sum_{(j,k) \in R_{ijk}} \beta_{ijk} \cdot \frac{P_{TT_{i}}}{|R_{ijk}|} \right) \right) \right]
}
\end{equation}

where \begin{itemize}
    \item $B_i$: is a binary variable that indicates whether railway service $i$ has been scheduled, represented as 1, or not scheduled, coded as 0. 
    \item $ca_i$: is the amount that the \gls{RU} operating service $i$ is willing to pay to the \gls{IM} for providing the service $i$ based on the requested times. Changes to the requested times will entail penalties in this value due to the inconvenience this causes to the \glspl{RU}.
    \item $P_{DT_{i}}$: is the maximum penalty applied for changes to the departure time for service $i$.
    \item $\alpha_{i}$: weights the penalty on the revenue due to deviation in the service's departure time with respect to the proposal of the \gls{RU} operating that service. This weight is obtained by the time difference between the \gls{RU}'s request and the allocation made by the \gls{IM}, as well as by the value of a constant for each \gls{RU} to determine the \gls{RU}'s sensitivity to schedule changes.
    \item $ R_{ijk} $ represents the ordered set of station pairs $ (j,k) $ that are visited sequentially by service $i$. Each pair $(j,k)$ indicates a direct travel segment between stations $j$ and $k$ along the route of service $i$. $|R_{ijk}|$ is the size of the set (pairs of stations of the service $i$).
    \item $P_{TT_{i}}$: is the maximum penalty applied for changes to the requested travel time between each pair of stations of the service $i$. The penalty percentage associated to each pair of stations of a service is obtained by dividing by the number of pairs $|R_{ijk}|$. Like $P_{DT}$, this value is also shared between all services.
    \item $\beta_{ijk}$: weights the penalty on the revenue applied due to deviation in travel time between a pair of stations $(j,k)$ visited by the service $i$. Travel times shorter than those requested are not allowed. This weight depends on the travel time deviation between pairs of stations, as well as on the value of the constant $k$ for each \gls{RU} to represent the \gls{RU}'s sensitivity to schedule changes.
    \item $ SS $: set of services, where $ |SS| $ is the number of services.
\end{itemize}

In summary, the objective function presented in Equation \ref{ecu:problem_formulation} reflects the \gls{IM}'s goal of maximizing profit by balancing revenue from scheduled services against penalties incurred due to deviations from the requested schedules. When the \glspl{RU} request their services from the \gls{IM}, these services have an associated amount that the \glspl{RU} are willing to pay for each service, provided that the service is planned with the same conditions as requested. In the modeling of the problem, it has been considered that any deviation from the original request will incur a penalty in the cost that the \gls{RU} is willing to pay for the service. Therefore, this is a decrease in the benefit to the \gls{IM} when it makes changes to the requests. The more variation, the greater the penalty. This model allows incorporating the case without penalties by setting these to zero by means of the variables $P_{{DT}_{{i}}}$ and $P_{TT_{ijk}}$. This balance ensures that only services meeting the operational constraints are scheduled, while changes to departure times or travel times are penalized according to the sensitivity of each \gls{RU}. 

Regarding the problem presented in Equation \ref{ecu:problem_formulation}, the following considerations should be taken into account: (1) In cases where all requests can be satisfied, the optimal solution will be to schedule all of them. However, this will not be the typical case in real-world deregulated scenarios, where conflicts will arise and must be resolved. (2) The modifications made to the original schedules during the optimization process must be evaluated to verify the feasibility of each service independently. Therefore, the conflict detection phase is crucial to ensure that the proposed solution is feasible. For example, in a system with two requests that present an unresolved conflict through schedule adjustments, only one of the two requests can be accepted. (3) The total profit obtained by the \gls{IM} does not always increase as the number of scheduled trains increases. For instance, consider a system with three conflicting service requests. In this case, it might be that only two of the requested services can be scheduled. At first glance, this seems to be the best solution. However, if the two scheduled services cover short-distance routes while the third serves a longer route, the profit from scheduling only the third request could be higher than that obtained from scheduling the other two.

\subsubsection{Problem constraints}
\label{constraints}

This section provides the formal definition of the problem constraints: departure time interval, operational times, dwell times and conflicts between services.

\textbf{Departure time interval:} The modified departure times ($  \tilde dt_{i}^{1} $) by the \gls{IM} must not exceed a specified margin $\delta$, either behind or in advance of the requested ones.


\begin{equation}
    dt_{i}^{1} - \delta \leq \tilde dt_{i}^{1} \leq dt_{i}^{1} + \delta
\end{equation}

where:
\begin{itemize}
  \item $dt_{i}^{j}$ is the original departure time at station $j$ of service $i$. In this first equation, only the departure time at the first station is considered.
  \item $\tilde dt_{i}^{j}$ is the updated departure time at the station $ j $.
  \item $\delta$ is the \gls{IM} modification margin.
\end{itemize}

\textbf{Operating time constraint:} the departure time proposed by the \gls{IM} for each station ($ \tilde dt_{i}^{j} $) must be greater than or equal to the departure time of the previous station ($ \tilde dt_{i}^{j-1} $) plus the original operating times (travel and dwell times). This constraint applies from the second until the penultimate stations. The first and last are excluded since the first lacks a previous station, and the last would not a have a departure time.


\begin{equation}
\tilde dt_{i}^{j} \geq \tilde dt_{i}^{j-1} + \tau_{i}^{j-1} + \sigma_{i}^{j}
\quad
(j=2,\dots, |R_{ijk}| - 1)
\end{equation}

where:
\begin{itemize}
  \item $ \tau_{i}^{j-1}$ is the travel time from stop $j-1$ to stop $j$ for service $i$.
  \item $\sigma_{i}^{j}$ is the minimal dwell time at stop $j-1$ for service $i$.
\end{itemize}

\textbf{Dwell time constraints: }the departure time from station $j$ must satisfy the following two restrictions:

\begin{enumerate}
    \item The departure time of station $ j $ ($  \tilde dt_{i}^{j} $) must be less than or equal to the sum of the original departure time ($ dt_{i}^{j} $) and maximum dwell time extension ($ \Delta_{\max} $).

    \begin{equation}
    \tilde dt_{i}^{j} \leq
    \,dt_{i}^{j} + \Delta_{\max} \quad
    (i=2,\dots,|R_{ijk}| - 1)
    \end{equation}

    \item The departure time from station $ j $ ($  \tilde dt_{i}^{j} $) must be less than or equal to the sum of the departure time from the previous station ($ \tilde dt_{i}^{j-1} $) plus operational travel time  ($ \tau_{i}^{j-1} $) plus maximum dwell time extension ($ \Delta_{\max} $).

    \begin{equation}
    \tilde dt_{i}^{j} \leq
    \tilde dt_{i}^{j-1} + \tau_{i}^{j-1} + \Delta_{\max}
    \quad
    (i=2,\dots,|R_{ijk}| - 1)
    \end{equation}
    
\end{enumerate}


where:
\begin{itemize}
  \item $at_{i}^{j}$ is the original arrival time at station $j$ of service $i$.
  \item $\Delta_{\max}$ is the maximum allowable dwell-time.
\end{itemize}

\textbf{Conflicts between services: } In order to identify any conflict between service $ i $ and service $ i' $, the values $ \Delta dt_{i,i'}^{j} $ and $ \Delta at_{i,i'}^{j+1} $ are calculated in the first instance (Equation \ref{eq:time_gaps_dt} and \ref{eq:time_gaps_at}). This values represent the time gap of the departure and arrival times between a common pair of stations for two services:


\begin{equation}
\Delta dt_{i,i'}^{j} \;=\; \tilde dt_{i}^{j} - \tilde dt_{i'}^{j}
\label{eq:time_gaps_dt}
\end{equation}

\begin{quote}
where $\Delta dt_{i,i'}^{j}$ is the departure‐time gap between service $i$ and $i'$ at station $j$.
\end{quote}

\begin{equation}
\Delta at_{i,i'}^{j+1} \;=\; \tilde at_{i}^{j+1} - \tilde at_{i'}^{j+1}
\label{eq:time_gaps_at}
\end{equation}

\begin{quote}
where $\Delta at_{i,i'}^{j+1}$ is the arrival‐time gap between service $i$ and $i'$ at station $j+1$.
\end{quote}

A conflict between two services arises when: the gaps have opposite signs (meaning that the two services intersect), the departure or arrival gap is less than twice the safety margin (because it is considered that each train carries its own safety margin). For example, considering two services $ i $ and $ i' $ with the following schedules between two shared stations:

\begin{itemize}
    \item Service $ i $: $ \tilde dt_{i}^{j} = 0 $ and $ \tilde at_{i}^{j+1} = 40 $.
    \item Service $ i' $: $ \tilde dt_{i'}^{j} = 5 $ and $ \tilde at_{i'}^{j+1} = 35 $.
\end{itemize}

Applying Equations  \ref{eq:time_gaps_dt} and \ref{eq:time_gaps_at} with the previous schedule times, the values $ \Delta dt_{i,i'}^{j} \;=\; \tilde dt_{i}^{j} - \tilde dt_{i'}^{j} = 0 - 5 = - 5 $ and $ \Delta at_{i,i'}^{j+1} \;=\; \tilde at_{i}^{j+1} - \tilde at_{i'}^{j+1} = 40 - 35 = 5 $ are obtained. This implies an overtaking of service  $ i' $ to $ i $ (service $ i' $ departed after service $ i $ but arrived earlier), which is not possible when considering a single track.

Formally, this constraint is represented as in Equation \ref{eq:conflicts}.


\begin{equation}
C_{i,i'} \;=\;
\begin{cases}
1, & 
\begin{aligned}[t]
&\exists\,(j,k)\in R_{ijk}\,\cap\,R_{i'jk},\;B_{i}=B_{i'}=1,\; \\
&\quad \Delta dt_{i,i'}^{\,j}\ \cdot \Delta at_{i,i'}^{\,j+1} \,\le\,0,\\
&\quad \bigl|\Delta dt_{i,i'}^{\,j}\bigr| < 2\ \cdot \omega,\\
&\quad \bigl|\Delta at_{i,i'}^{\,j+1}\bigr| < 2\ \cdot \omega;
\end{aligned}
\\[2ex]
0, & \text{otherwise.}
\end{cases}
\label{eq:conflicts}
\end{equation}

where:
\begin{itemize}
  \item $C_{i,i'}$ is the conflict indicator between services $i$ and $i'$ ($1=$ conflict, $0=$ no conflict).
\end{itemize}

\subsection{Modular system designed}
\label{sec:modular_system_designed}

\begin{figure}[ht]
    \centering
    \includegraphics[width=0.45\textwidth]{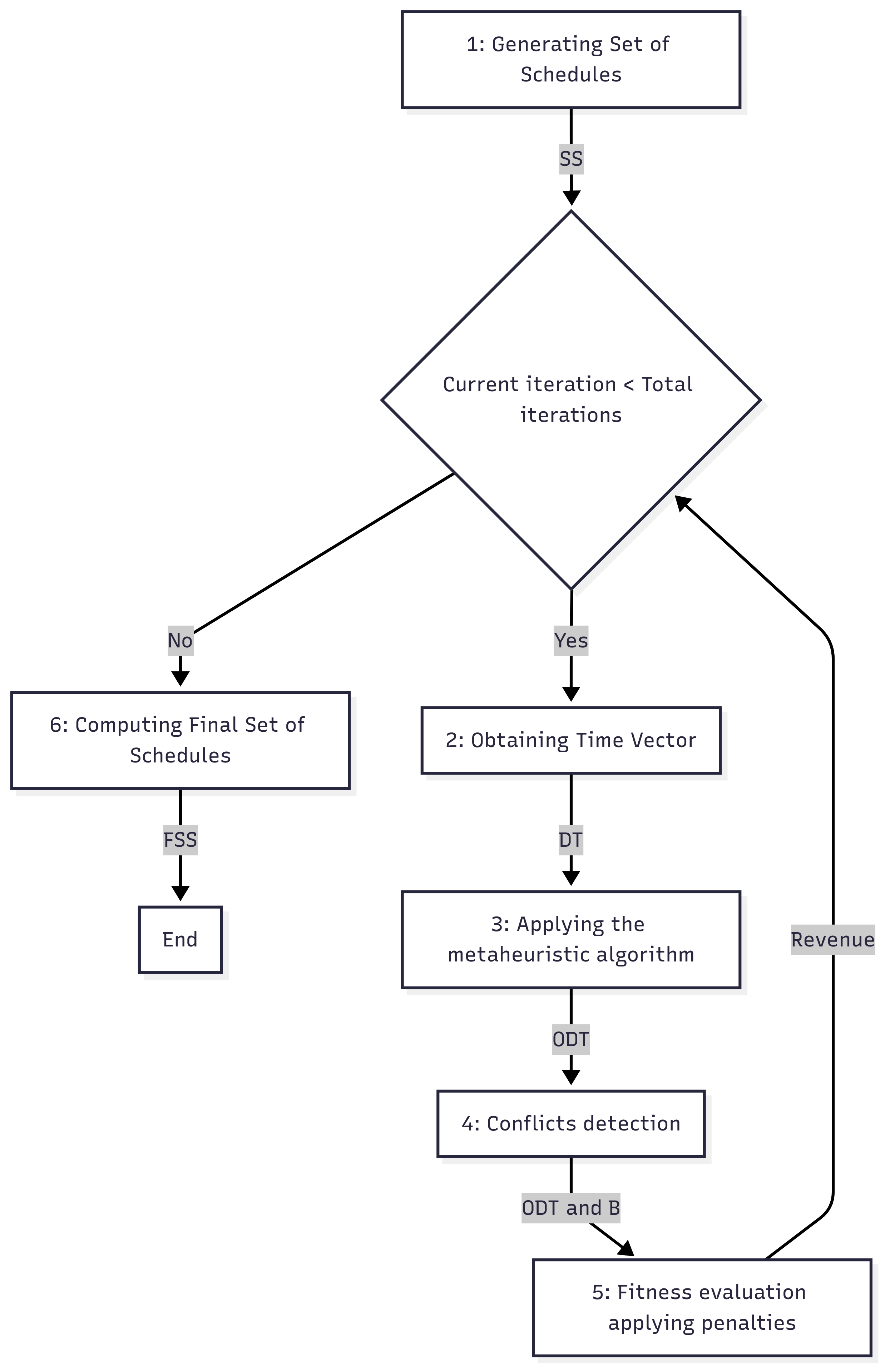}
    \caption{Flowchart of the designed modular system.}
    \label{fig:modular_system}
\end{figure}

A modular system has been developed to perform simulations and test different optimization methods, as shown in the flowchart in Figure \ref{fig:modular_system}. It is important to highlight that the diagram names the structures, although they are formally defined in Section \ref{sec:general_scheme}. The first step consists of obtaining the schedule proposals from the RUs. This is done using a module of the ROBIN simulator that generates the structure $SS$ (Set of Schedules), which contains the proposals from each \gls{RU} (Section \ref{sec:generation_module}). Next, the iterations of the metaheuristic algorithm begin, where Steps 2 to 5 occur in each iteration. $SS$ is transformed into the departure time vector $DT$, suitable as input for the metaheuristic algorithm, which produces the optimized departure times $ODT$ as output. $ODT$ may lead to conflicts between services, which are resolved in Step 4, resulting in a conflict-free vector $ODT$ and the vector $B$ from Equation \ref{ecu:problem_formulation} — the general equation of the formulation — which indicates the services that would be scheduled. With these values, the revenue is calculated by applying the penalties based on the changes made in Step 4. Then, it is checked whether a new iteration is necessary. Once all iterations have been completed, the final service assignment is obtained in $FSS$.

What follows details how the generation of schedule requests (Step 1, Section \ref{sec:generation_module}), conflict detection (Step 4, Section \ref{sec:conflict_detection_module}), and penalty calculation (Step 5, Section \ref{sec:penalty_calculation_module}) are carried out.

\subsubsection{Generation module}
\label{sec:generation_module}

This module generates a set of requests for services addressed to the \gls{IM} in order to obtain the reservation of the tracks necessary to meet the timetables desired by each \gls{RU}. The requests detail the timetable for the use of each service, specifically, for each request, the stations to be visited and the arrival and departure times at each of these stations are indicated. It is a module taken from the ROBIN \cite{delCastillo24} simulator which generates timetable requests in a pseudo-random way. It's considered that each \gls{RU} submits only as many service requests as its allocated infrastructure can support—namely, the total number of trains that its assigned storage and maintenance yards can accommodate. In other words, the model does not permit an unbounded number of services: the input request set for each \gls{RU} has been pre‐filtered so as not to exceed its yard capacity. One important aspect is the generation of the $ ca_i $ value. As seen in Equation \ref{ecu:problem_formulation}, this value relates to the willingness of the \gls{RU} to pay to schedule the requested service, and the associated penalties due to any changes in the original schedule are also related to this value. In this study, $ ca_i $ is calculated for each service based on two questions: the number of stopsthe service makes and the \gls{RU} operating it. This approach recognizes that different \glspl{RU} may have varying levels of financial capacity or strategic interest, justifying the adjustment of $ ca_i $ to reflect these variations in a reasonable manner.

\subsubsection{Conflict detection module}
\label{sec:conflict_detection_module}

This module is responsible for detecting conflicts in the timetable requests received from the \glspl{RU}. To perform this function effectively, a method is required to ensure quick conflict detection. 

\begin{table}[ht]
\centering
\caption{Timetables of rail services organized by station.}
\begin{tabular}{@{}cccc@{}}
\toprule
Station   & \multicolumn{3}{c}{Service ID} \\ \midrule
           & Service 1 & Service 2  & Service 3 \\ \midrule
Madrid     & (18:20, 18:20) &      -       & (18:00, 18:00)  \\
Calatayud  &       -       &      -       & (18:50, 18:54)  \\
Zaragoza   &       -       & (19:50, 19:50) &       -        \\
Lleida     & (19:55, 19:55) &       -      & (20:10, 20:14)  \\
Tarragona  &       -       &      -       &       -        \\
Barcelona  &       -       & (21:00, 21:00) & (21:20, 21:20)  \\ \bottomrule
\end{tabular}
\label{tab:servicios_ferroviarios}
\end{table}

\begin{figure}[ht]
    \centering
    \includegraphics[width=0.5\textwidth]{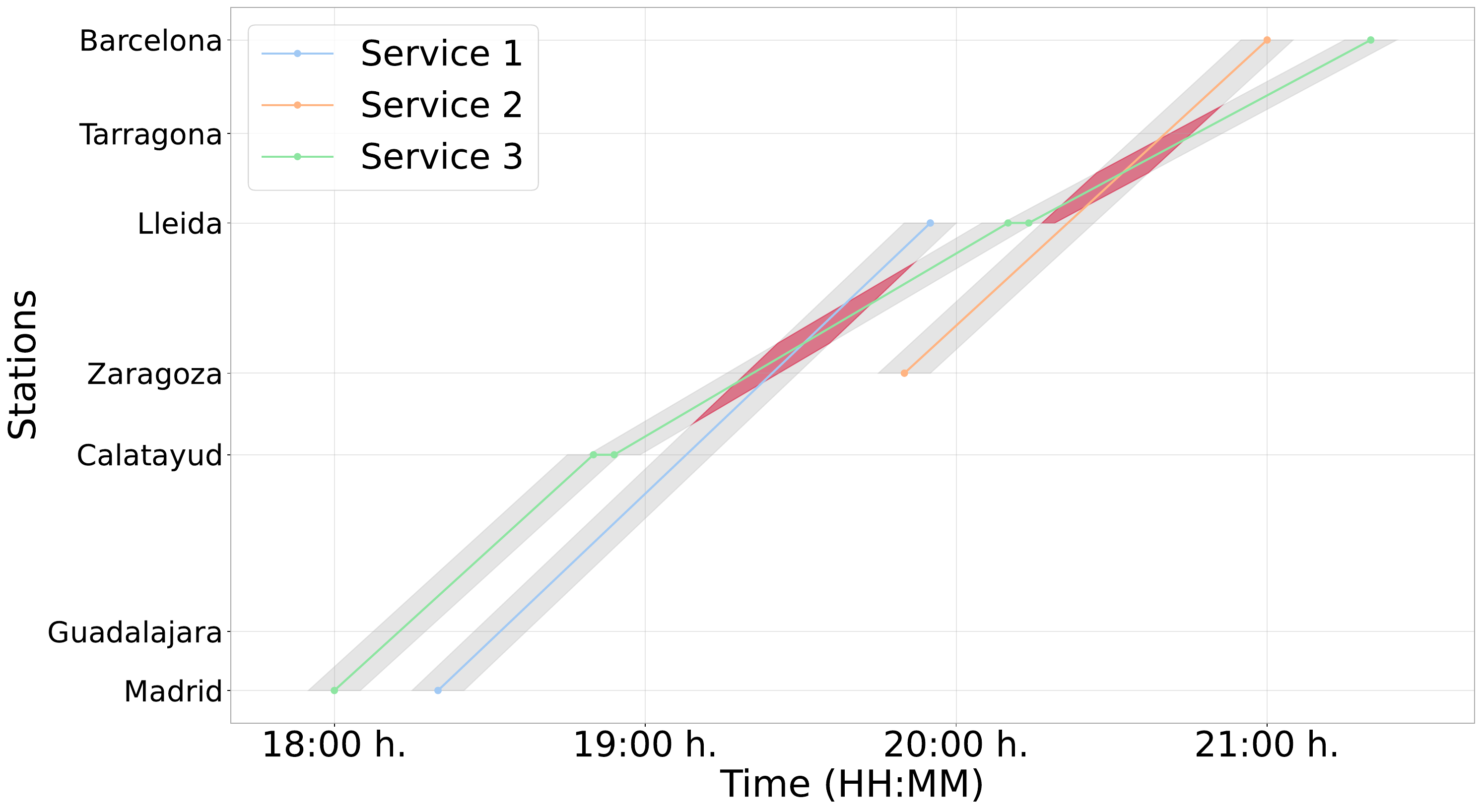}
    \caption{Marey chart representation of the timetables.}
    \label{fig:marey_example} 
\end{figure}

Table \ref{tab:servicios_ferroviarios} shows the timetables of three services on the Madrid-Barcelona line. Considering that each tuple of values represents the arrival time ($t_a^{STA}$) of a train at the station and the departure time ($t_d^{STA}$). For instance, Service three initially departs Madrid at $t_d^{MAD} = 18:00h$, making a four-minute stop at $t_a^{CAL} = 18:50h $ in Calatayud. Subsequently, it arrives in Lleida at $t_a^{LLE} = 20:10h$, where it makes a new four-minute stop. Finally, it arrives in Barcelona (end of the trip) at $t_a^{BAR} = 21:20h$. Figure \ref{fig:marey_example} shows the corresponding Marey chart for the three services presented in Table \ref{tab:servicios_ferroviarios}. Conflicts between services are highlighted in red, considering a 10 minute safety headway for each service.

The representation of conflicts between rail services is an important aspect in the modeling of the problem. In this case, a matrix structure has been used to represent them. Thus, the conflicts between a set of services of order $n$ can be represented by means of a matrix $C_{nxn}$ whose dimension is equal to the number of services. $C$ is therefore a square and symmetric matrix, since each of its elements $c_{ij}$ represents the presence or absence of conflicts between services $i$ and $j$. For this reason, it is a binary matrix, whose values are 1 (in the case of conflict) or 0 (when there is no conflict). The following matrix therefore allows the existence of conflicts between services 1 and 3, as well as between 2 and 3 (and vice versa) to be reflected.


\begin{equation}
 C = \begin{pmatrix}
  0 & 0 & 1 \\ 
  0 & 0 & 1 \\ 
  1 & 1 & 0 \\ 
\end{pmatrix}
\label{eq:mconflicts}
\end{equation}

The conflict representation method presented in this paper assumes that a conflict arises when, at any time over the course of a service, there is an intersection with the safety area of another service. However, considering the motion of trains as a \gls{URM}, it is possible to simplify the search for conflicts. At the strategic planning stage—i.e., the process of developing long‐range, year-ahead timetables and resource allocations several months before service deployment—a \gls{URM} approximation in which each train’s between-stop trajectory is modeled as a constant-speed segment has been considered. This simplification reduces model complexity and enables efficient, network-wide conflict detection. To absorb unmodeled variability in running speeds, acceleration/deceleration profiles and station dwell times, a fixed safety headway $\omega$ has been embedded into the conflict-gap constraints—thereby reserving extra time-slots around each passage and preventing spurious conflicts even when actual operations deviate from the nominal \gls{URM} profile. Then, if the start time of service \(i'\) is later than the end time of service \(i\), no conflict exists, and the check concludes quickly. In other cases, it examines all relevant stations, including shared stations and those along the route of service \(i\). At each station, conflicts are evaluated based on three conditions: the gap between the departure times of the two trains at the initial station is smaller than the safety margin; the gap between their arrival times at the final station is smaller than the safety margin; and even whether the time gaps are sufficient, a conflict arises if their signs differ, indicating that the paths of the two trains intersect at some point along the route. By assuming that the trains' trajectories follow a \gls{URM},  it is not necessary for service $i'$ to visit exactly the stations of the path being checked to obtain the time value at each station; these values can be inferred from the equation of a line given two points.

\subsubsection{Penalty calculation module}
\label{sec:penalty_calculation_module}

The objective of the IM, as previously mentioned, is to maximize its profit. To do so, given a series of requests that may present conflicts, the \gls{IM} will generate a feasible timetable in which it will plan a series of services from all the requests received. The profit associated with the final schedule depends on the access fee that each \gls{RU} pays to the \gls{IM} for the use of the infrastructure. However, the schedule adjustments that the \gls{IM} makes to the received requests result in penalties on this fee, reducing its profit. Therefore, the greater the deviation planned by the \gls{IM} for a service's schedule from the initially submitted proposal, the higher the penalty. This penalty on the fee affects, on the one hand, the deviation in the departure time of the service and, on the other, the sum of each deviation in departure times at all stations the service visits along its route. The timetabling problem addressed in this study is an annual, in-advance planning exercise, carried out several months before actual service deployment. Consequently, all market conditions used to calibrate penalty coefficients are those forecast by the respective \glspl{RU} at the start of the planning cycle, and no subsequent updates to these estimates occur during the optimization. In other words, slot allocations—and hence penalty values—are based on the static, \gls{RU}-provided forecasts, so time-dependent variation in penalty parameters has not been considered. Suppose a train travels from Madrid to Calatayud to Zaragoza to Barcelona. The penalty associated with the fee paid by the \gls{RU} for this service is related to the deviation that the \gls{IM} has planned from the departure time of this train compared to the proposal received from the \gls{RU}. Additionally, it also depends on the sum of the deviations in departure times at each station visited along the route, in this case, Calatayud and Zaragoza. This penalty will be different for each RU, i.e., each \gls{RU} will be affected differently. It will be modeled with what has been called the ``penalty sensitivity curve''.

To this end, another module has been developed that takes as input the service requests obtained earlier. For each request, this module allows a pattern to be generated representing how the IM's schedule will impact the profit obtained by scheduling that request, considering the type of service and the \gls{RU} operating it. This pattern associated with each service request is generated based on the following criteria:

\begin{itemize}
    \item Let there be an access fee to the infrastructure $ca$. The value of this fee is directly proportional to the number of stations visited by a given service. Thus, the greater the number of stations visited by a service, the higher the access fee paid by the \gls{RU} to operate it, and the higher the profit obtained by the \gls{IM}.
    \item A random factor is introduced to add variability between different services visiting the same number of stations.
    \item The parameter $P_{max}$ is defined as the maximum penalty value imposed on the fee. In the tests carried out in the paper, $P_{max} = 0.4$, meaning that if a service time proposal made by \gls{IM} varies up to the maximum allowed interval from the request made by the \gls{RU}, the \gls{IM} will receive 40\% less revenue from the infrastructure access fee for that service.
    \item The parameter $P_{DT_{i}}$ is introduced as the maximum penalty for changes in the service's departure time. In the tests reported in this study, $P_{DT_{i}} = 0.35$, which means that a maximum of 35\% of the penalty imposed on a service fee is due to changes in its departure time.
    \item The parameter $P_{TT_{i}}$ is introduced as the penalty on the access fee associated with changes in the travel times of the service at its stations visited. This value is divided among all the origin-destination pairs of the service, so if it is a direct service (no intermediate stops), the entire penalty on the fee is due to the deviation in the train's departure time. In this study, $P_{TT_{i}}=0.65$.
    \item The constant $ k $ models the penalty sensitivity curve of an \gls{RU} based on the schedule changes made by the \gls{IM}. The penalty sensitivity curve is intended to model the effect on a \gls{RU} of changes in schedules by the \gls{IM}; i.e. each \gls{RU} will be more or less sensitive to these as a function of this curve defined by $k$.
    \end{itemize}

The values of $P_{max}$, $P_{DT_{i}}$, $P_{TT_{i}}$ are configurable in the model and have been adjusted empirically.

To model the sensitivity to the penalty, Equation \ref{ecu:penalty_curve} is used. The selected penalty curve tries to model how modifications on the requested times will affect the willingness to pay, considering that the modifications proposed by the \gls{IM} are always within a margin (e.g. 60 minutes). If there are no changes to the request, no penalties are incurred. On the other hand, as the modification proposed by the \gls{IM} approaches the maximum margin, the corresponding maximum penalty is incurred. In addition, the curve tries to model different types of sensitivity based on the value of parameter k, so that the willingness to pay of the \glspl{RU} is not always affected in the same way.

\begin{equation} \label{ecu:penalty_curve}
f(x,k) = 1 - e^{-k x^2} \left(\frac{1}{2} \cos(\pi x) + \frac{1}{2}\right)
\end{equation}

where $k$ is the variable that modulates the sensitivity of the curve, and $x$ is the relative displacement between the requested service times made by a {RU} and the times proposed or modified by the \gls{IM}. This time difference is normalized to the interval $[0,1]$ (zero if there is no difference between requested times and the \gls{IM} proposal, and one if the difference between them is the maximum allowed). Differences above the constraint will result in an unfeasible service. To achieve this value, the time difference is divided by the maximum displacement, which in this case corresponds to the value of the safety margin.

\begin{figure}[ht]
    \centering\includegraphics[width=0.45\textwidth]{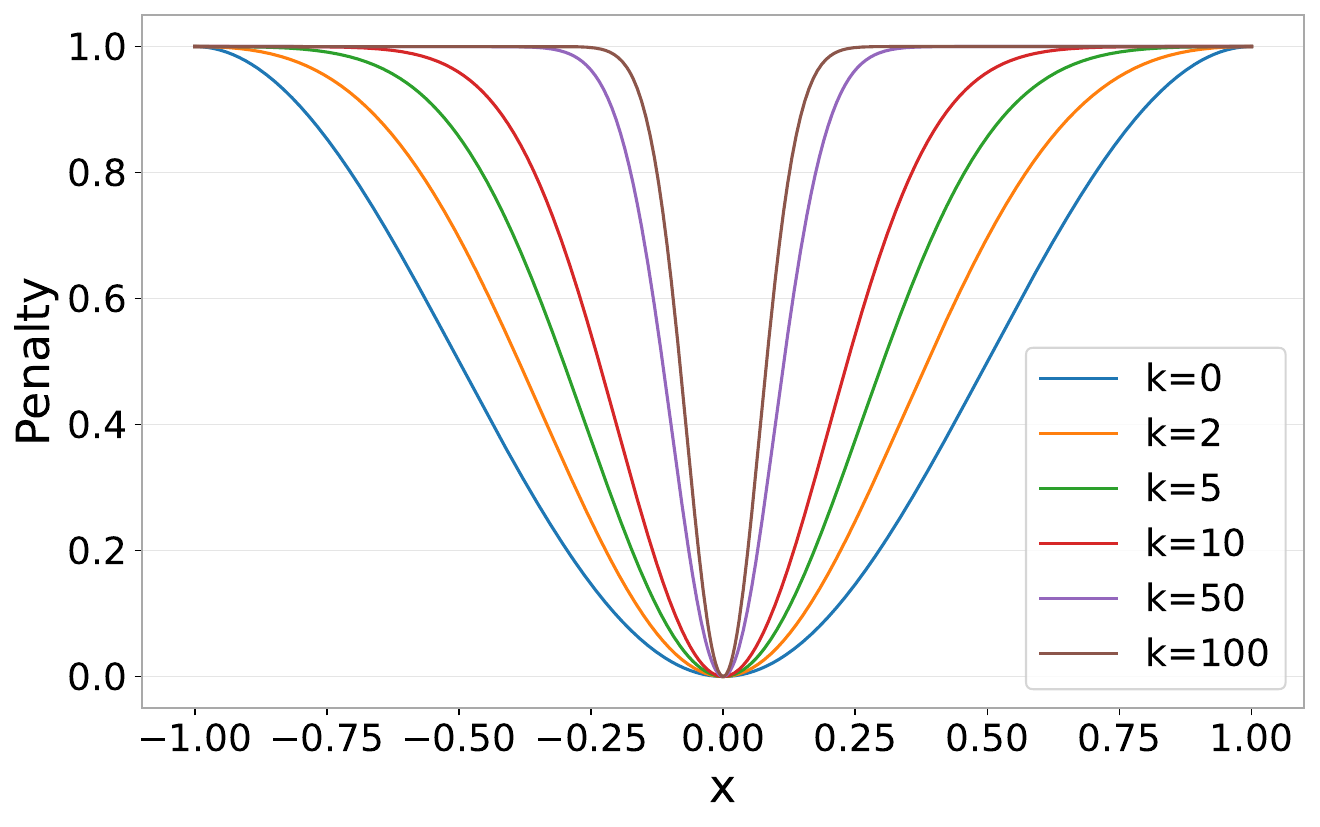}
    \caption{Penalty sensitivity curve.}
    \label{fig:penalty} 
\end{figure}

Figure \ref{fig:penalty} represents the effect of the value of the constant $k$ on the curve. As can be observed, the higher the value of the constant $k$, the greater the sensitivity to schedule changes, resulting in higher penalty values.

In order to calculate the revenue obtained with an \gls{IM} proposal, a maximum penalty is considered for changes in the service departure times ($ P_{DT_{i}} $) and changes in each one of the travel times ($P_{TT_{ijk}}$) for each service $i$. For example, for a maximum deviation of 10 minutes between proposals and requests, if no modification is made with respect to the requested departure time, the difference would be zero ($ x = 0$). In this case, the penalty associated with changes in departure time will be multiplied by this value (i.e. no penalty would be applied). On the other hand, if the proposed departure time is modified by 10 minutes with respect to that requested, the maximum departure time penalty will be applied.

$\alpha_i$ is the penalty percentage shown in Equation \ref{ecu:problem_formulation} for the \gls{RU} $i$. The value of $ \alpha_i$ is obtained by using Equation \ref{ecu:alpha_X_DT}, where the value of the constant $k_i$ depends on the \gls{RU} $i$ operating the service (Equation \ref{ecu:alpha_X_DT}).
\begin{equation}
    \label{ecu:alpha_X_DT}
    \alpha_i =  |f(x_{DT}, k_i)|
\end{equation}
\noindent where $ 0 \leq \alpha_i \leq 1 $.

To obtain the penalties based on a new \gls{IM} proposed times (i.e. solution obtained with a metaheuristic algorithm), the first step is to calculate the relative difference between the requested and proposed departure times $ x_{DT} $ (Equation \ref{ecu:X_DT}).
\begin{equation}\label{ecu:X_DT}
    x_{DT} = \frac{IM_{DT} - RU_{DT} }{\delta}
\end{equation}
\noindent where $ IM_{DT} $ is the \text{\gls{IM} proposed departure time}, $ RU_{DT} $ is the \text{\gls{RU} requested departure time} and $ \delta $ defines the maximum interval in which the requested travel times can be modified. This study considers a maximum of 10 minutes (i.e. service departure times can be moved back or forward a maximum of 10 minutes).

Next, the penalties associated with travel time changes are obtained using the previous approach. The process is analogous to the previous one based on Equation \ref{ecu:beta_ijk}.
\begin{equation}\label{ecu:beta_ijk}
    \beta_{ijk} =  |f(x_{TT}, k_i)|
\end{equation}
\noindent where $ 0 \leq \beta_{ijk} \leq 1 $.

\begin{equation}\label{ecu:X_TT}
    x_{TT} = \frac{IM_{TT} - RU_{TT}}{\delta}
\end{equation}
\noindent where $IM_{TT}$ is the \gls{IM} proposed travel time between a pair of stations, $RU_{TT}$ the \gls{RU} requested travel time between a pair of stations and  $ \delta $ defines the maximum interval in which the requested departure times can be modified (10 minutes).

The computation of $\beta_{ijk}$ is based on calculating the relative difference between the requested and proposed travel times. This process is repeated for every trip (origin-destination travel) of the service.

\subsection{General scheme of the optimization process}
\label{sec:general_scheme}

\begin{figure}[ht]
    \centering
    \includegraphics[width=0.49\textwidth]{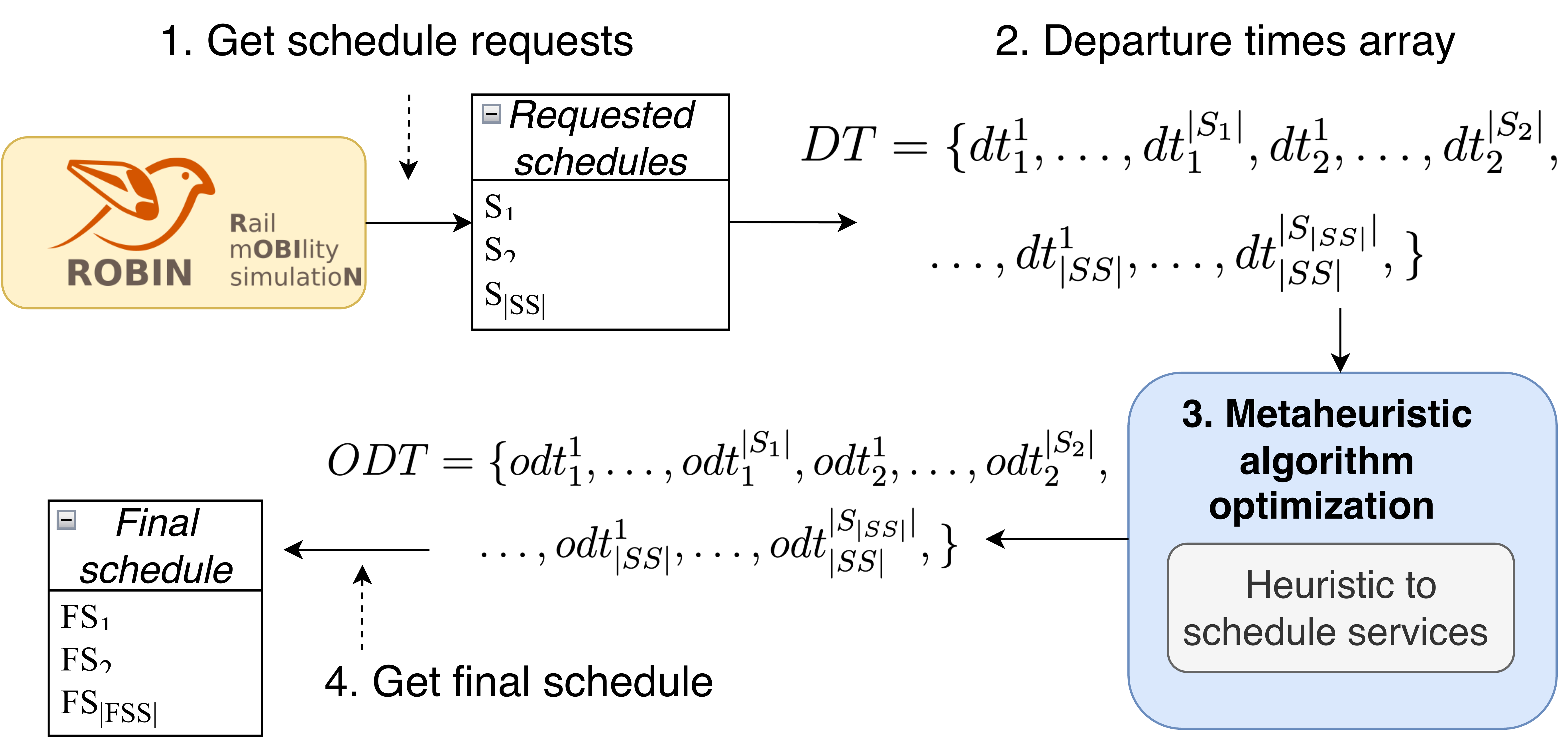}
    \caption{General optimization scheme.}
    \label{fig:optimization_scheme} 
\end{figure}

The general scheme of the optimization process is shown in Figure \ref{fig:optimization_scheme}. Each of the steps involved in this process will be described in detail, and a step-by-step example will also be shown to clarify how the optimization process works.

\noindent \textbf{1. Generating schedule requests:} 

    Let $SS = \{ S_1, \dots, S_{|SS|} \}$ be the set of schedules, where each $S_i$ is a schedule. Each service in $S_i$ is composed of a set of stops. For each stop, ROBIN indicates the station ($Sta_i^j$), arrival time ($at_i^j$) and departure ($dt_i^j$). Thus, $ S_i $ can be expressed as:
    \begin{equation}\label{lab:set_of_schedule}
            S_i = \{ (Sta_i^1,at_i^1,dt_i^1), \dots,(Sta_i^{|S_i|},at_i^{|S_i|},dt_i^{|S_i|})\}
    \end{equation}
    
    where the operator $|\ |$ denotes the cardinal of the set it refers to.
    
    For example, let the following timetables be associated with three rail services as shown in Table \ref{tab:servicios_ferroviarios}. $SS$ will be defined as $SS = \{ S_1, S_2, S_3 \}$ where for instance  $S_3$ is
    \{ (Madrid, 18:00, 18:00), (Calatayud, 18:50, 18:54),$\\(Lleida, 20:10, 20:14),(Barcelona, 21:20, 21:20)\}$

\noindent \textbf{2. Computing departure time vector:} The metaheuristic algorithms will receive a vector with the departure times as input, considering that the original operational times must be complied with (stop time plus travel time). For this purpose, a one-dimension vector containing the departure times of the services in a sorted way is generated using $SS$. This vector is called $DT$ and can be expressed as:
    
    \begin{equation}\label{ecu:departure_times}
        \begin{aligned}
        DT = \{ & dt_1^1,\dots,dt_1^{|S_1|},
        dt_2^1,\dots,dt_2^{|S_2|},\\
         & \dots, dt_{|SS|}^1,\dots,dt_{|SS|}^{|S_{|SS|}|},\}
        \end{aligned}
    \end{equation}

    Based on the previous example schedule, the departure times vector will be as follows: DT = [18:20, 19:55, 19:50, 21:00, 18:00, 18:50, 20:10, 21:20]
    
\noindent \textbf{3. Optimizing the departure time vector:} The metaheuristic algorithms take the previous vector as input, and return the Optimized Departure Time ($ODT$). The $DT$ vector is optimized by the algorithm; by adjusting departure times, more services could be scheduled compared with the original requests, maximizing the \gls{IM} revenue.

The metaheuristic algorithms tested take $DT$ as input and return $ODT$ as output, which contains the optimized times. $ ODT $ can be written as follows:
    
\begin{equation}\label{ecu:opt_departure_times}
\begin{aligned}
    ODT = \{ & odt_1^1, \dots, odt_1^{|S_1|}, odt_2^1, \dots, odt_2^{|S_2|}, \\
             & \dots, odt_{|SS|}^1, \dots, odt_{|SS|}^{|S_{|SS|}|} \}
\end{aligned}
\end{equation}
    
    As can be observed, $ODT$ has the same format as $DT$, while $ ODT $ contains the optimized values of the departure times obtained by the metaheuristic algorithm tested.

    In the example, the metaheuristic algorithm can reach solutions such as the following: 
    
    $ODT = [\mathbf{17:50}, \mathbf{19:25}, \mathbf{20:20}, \mathbf{21:30}, 18:00, 18:50, 20:10, 21:20]$

    As mentioned above, with the original requested times, it was not feasible to schedule all three services. However, based on the new $ODT$ times this would be feasible, if a five minute minimum safety gap for each service is included. In order to reach this solution, $S_1$ has been advanced by half an hour, and the departure of Service 2 has been delayed, also by thirty minutes.        

    It is important to remark that this might not be the optimal solution: for example, if \gls{RU} running service one is much more sensitive to penalties compared to the others, it would be better to delay Service 3 and Service 3 while not modifying the requested times of Service 1.
    
\noindent  \textbf{4. Construction of the final schedule:} The final schedule $FSS$ is obtained based on the $ODT$ achieved by the metaheuristic, where $FS_i$ is the final schedule for each service $S_i$. In order to achieve a feasible schedule, the heuristic shown in Algorithm \ref{alg:heuristico_planificacion} is used. This heuristic is detailed in Section \ref{ecu:a_heuristic_for_railway_service_scheduling}. The final schedule $FSS$ follows the expression below:
    \begin{equation}\label{ecu:final_set_of_schedule}
        FSS = \{ FS_1, \dots, FS_{|FSS|} \}
    \end{equation}
    
\begin{figure}[ht]
    \centering
    \includegraphics[width=0.5\textwidth]{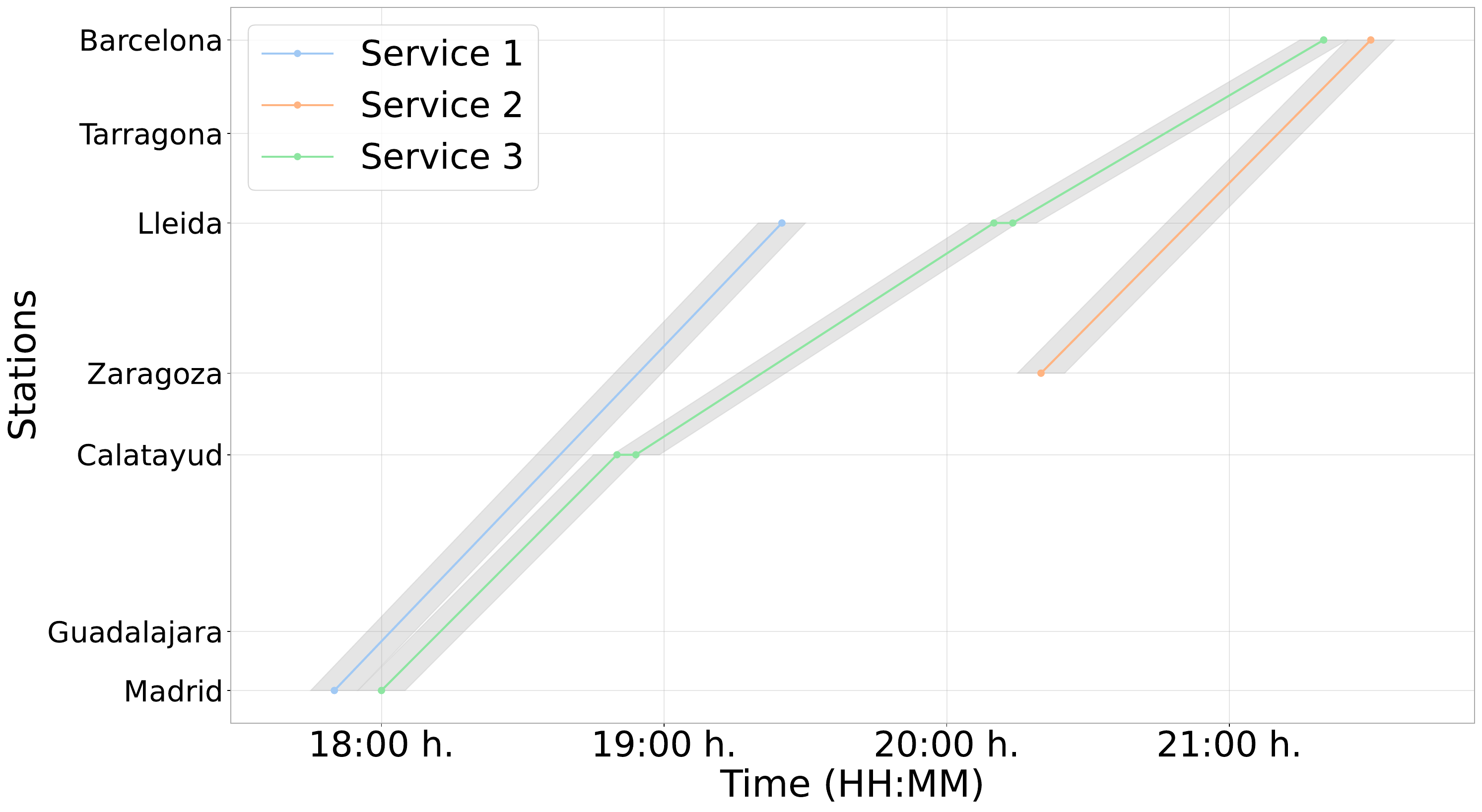}
    \caption{Mary chart schedule based on $ODT$.}
    \label{fig:marey_example_optimized_gaps} 
\end{figure}

    In the following example, based on the previous $ ODT $ vector, a definitive schedule is devised. Even with the optimized departure time vector, it is possible to find conflicts between services, making it infeasible to schedule all the requests. For this reason, the heuristic for service planning is also applied. The Marey chart with the final schedule based on the $ ODT $ previously shown for the example schedule can be seen in Figure \ref{fig:marey_example_optimized_gaps}. The modified schedule resulting from the $ ODT $ vector is set out in Table \ref{tab:servicios_ajustados}.

\begin{table}[ht]
\centering
\caption{Final timetable of rail services organized by station.}
\begin{tabular}{@{}cccc@{}}
\toprule
Station   & \multicolumn{3}{c}{Service ID} \\ \midrule
           & Service 1 & Service 2  & Service 3 \\ \midrule
Madrid     & (17:50, 17:50) &      -       & (18:00, 18:00)  \\
Calatayud  &       -       &      -       & (18:50, 18:54)  \\
Zaragoza   &       -       & (20:20, 20:20) &       -        \\
Lleida     & (19:25, 19:25) &       -      & (20:10, 20:14)  \\
Tarragona  &       -       &      -       &       -        \\
Barcelona  &       -       & (21:30, 21:30) & (21:20, 21:20)  \\ \bottomrule
\end{tabular}
\label{tab:servicios_ajustados}
\end{table}

\subsubsection{A heuristic for railway service scheduling}
\label{ecu:a_heuristic_for_railway_service_scheduling}

The optimization process carried out by the metaheuristic algorithms is exclusively responsible for modifying the vector with departure times. Therefore, the output of the algorithm does not correspond to a service schedule based on the requests received. Thus, the following questions remain to be answered: How can it be determined which railway services will be scheduled with the times obtained by the algorithms? Which feasible combination will provide the greatest benefit to the manager?

Feasible train combinations are those where none of the scheduled services present conflicts with any others. Each service will thus have two possible states: scheduled or not scheduled. In this way, it is possible to represent the sequence of scheduled trains using a vector of Boolean values, where a true value indicates that the service has been scheduled, and false otherwise. Based on this, the number of possible combinations to be checked increases as the number of services grows, on the order of $2^n$. This exponential growth makes it unfeasible to check all possible combinations for a sufficiently large number of services.

To address the exponential growth caused by checking all possible combinations, the following heuristic technique has been implemented to obtain the vector of trains to be scheduled. The operation of this heuristic is detailed in Algorithm \ref{alg:heuristico_planificacion}. This algorithm aims to schedule a set of services represented by the $ODT$ vector, resolving conflicts between them to produce a final conflict-free scheduled services vector called $B$. Initially, conflicts between the proposed services are identified (Line 3), and those without conflicts are scheduled (Line 4). Then, the benefits associated with the conflicting services are calculated (Line 5). As long as there are services in conflict (Line 6), the algorithm iteratively selects the service with the highest associated benefit (Line 7), schedules it (Line 8), and updates the list of conflicts by removing the newly scheduled services and their corresponding conflicts (Line 9). This approach seeks to maximize total benefits by prioritizing high-value services and minimizing interference.


\begin{algorithm}
\caption{Heuristic for obtaining the services to be scheduled}
\label{alg:heuristico_planificacion}
\begin{algorithmic}[1]
\STATE \textbf{Input:} ODT (Proposed time vector)
\STATE \textbf{Output:} B (Scheduled services vector)
\STATE C = CONFLICTS\_IN(ODT) 
\STATE B = SCHEDULE\_NO\_CONFLICTS\_IN(C) 
\STATE R = REVENUES\_OF(C) 
\WHILE{C $\neq \emptyset$}
\STATE S = MAX(R) 
\STATE B = B + S 
\STATE C = C - CONFLICTS\_BETWEEN(B, S) 
\ENDWHILE
\end{algorithmic}
\end{algorithm}

It is important to underline the deterministic nature of the heuristic for service scheduling, in reaching the same service scheduling with different evaluations of the same values with the $ DT $ or $ ODT $ vectors.

\section{Experiments and results}
\label{sec:results}

This section presents the results of metaheuristic algorithms for solving the timetabling problem in a liberalized railway market. Section \ref{sec:sensitivity_analysis} analyzes the impact of the selected safety headway and \gls{IM} bound in the obtained results and provides further insights on the selected values. Section \ref{sec:tests_without_hiperparameter_tunning} shows the results obtained without hyperparameter optimization, and then after optimization (Section \ref{sec:tests_witht_hiperparameter_tunning}). This enables evaluation of their effectiveness and configuration recommendations. The Python library used is \textit{MEALPY} \cite{Vanthieu2023}, as it is  robust and efficient and implements a large number of cutting-edge population-based metaheuristic algorithms. Finally, Section \ref{sec:problem_scalability} shows the results obtained by the metaheristic algorithms in a larger scale instance with 50 service requests, and compares the results with those obtained by classical mathematical approaches such as the \gls{SCIP} solver.

\subsection{Sensitivity analysis between safety margin \& \gls{IM} bounds}
\label{sec:sensitivity_analysis}

\begin{figure}[ht]
    \centering
    \includegraphics[width=0.5\textwidth]{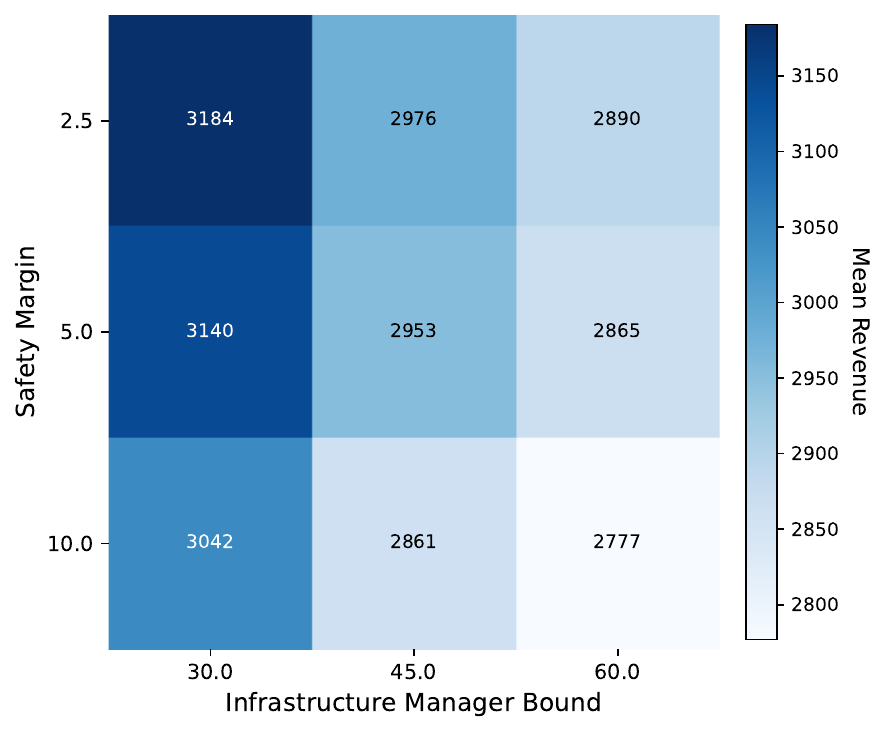}
    \caption{Revenue sensitivity to safety margins and infrastructure bounds.}
    \label{fig:general_sensitiviy_heatmap} 
\end{figure}

In this subsection, a sensitivity analysis will be carried out regarding the safety margin ($\omega$) and \gls{IM} bound ($\delta$) impact to the revenue performance. Specifically, 2.5, 5, and 10 for safety margins $\delta$, and 30, 45, and 60 minutes for \gls{IM} bounds $\omega$ has been tested. These combinations allow to observe how increasing the buffer for train spacing and the \gls{IM} modification margin affect average revenue.
Figure~\ref{fig:general_sensitiviy_heatmap} presents a heatmap of mean revenue as a function of safety margin (rows) and \gls{IM} bound (columns).  The color gradient clearly indicates that the highest average revenue (darkest blue) occurs when the safety margin is 2.5 minutes and the \gls{IM} bound is 30 minutes, while the lowest occurs at a 10 minute safety margin with a 60 minute \gls{IM} bound.  In general, revenue tends to decrease as either safety margin or \gls{IM} bound increases.

An ANOVA test has been carried out with the revenue obtained in five different runs of each meataheuristic with each combination of $\delta$ and $\omega$ (450 values), with the objective of analyzing the impact of each parameter both individually and combined. The two-way ANOVA on mean revenue shows that the \gls{IM} bound has a statistically significant effect on revenue (\(F(2,\,441)=4.32,\;p=0.0139\)), while the safety margin does not (\(F(2,\,441)=0.88,\;p=0.414\)), and there is no significant interaction between them (\(F(4,\,441)=0.0050,\;p=0.9999\)).  The overall regression model explains only about 2.3 \% of the variance (\(R^2=0.023\)), indicating that most of the variability in revenue is due to factors beyond these two experimental parameters.

When examined by individual algorithm, the following patterns emerge:
\begin{itemize}
    \item \textbf{Genetic Algorithm}: Safety margin has a highly significant effect (\(p<0.0001\)), and \gls{IM} bound is also significant (\(p=0.030\)). No interaction effect is observed.
    \item \textbf{Particle Swarm Optimization}: Safety margin remains significant (\(p=0.0015\)), whereas \gls{IM} bound does not (\(p=0.293\)). Interaction is non-significant.
    \item \textbf{Differential Evolution}: Neither safety nor infrastructure factors are significant (\(p>0.05\) in all cases).
    \item \textbf{CMA-ES}: \gls{IM} bound is extremely significant (\(p<10^{-7}\)), while safety is marginal (\(p\approx0.0629\)), and no interaction effect is detected.
    \item \textbf{Whale Optimization Algorithm}: \gls{IM} bound shows a significant effect (\(p=0.0082\)), safety margin does not (\(p=0.870\)), and there is no interaction.
    \item \textbf{Artificial Bee Colony}: Both safety margin (\(p=0.007\)) and infrastructure bound (\(p<10^{-7}\)) are significant; interaction is non-significant.
    \item \textbf{Grey Wolf Optimizer}: Only \gls{IM} bound is significant (\(p=0.0091\)); safety and interaction are not.
    \item \textbf{Ant Colony Optimization (continuous)}: \gls{IM} bound is highly significant (\(p<10^{-8}\)); safety and interaction are not.
    \item \textbf{Simulated Annealing}: Neither safety nor infrastructure factors reach significance (\(p>0.05\)).
    \item \textbf{Hybrid GWO-WOA}: Safety margin is significant (\(p=0.0199\)) and there is a significant safety × infrastructure interaction (\(p=0.0146\)); \gls{IM} bound alone is not significant in this hybrid scheme.
\end{itemize}

In all experiments, a 10 minute safety margin was chosen for operational and safety reasons (allowing sufficient buffer to absorb minor delays without violating headway constraints). The 60 minute modification margin for the \gls{IM} corresponds to the value that ADIF (the Spanish \gls{IM}) states as reasonable in its official declaration on network operations \cite{Adif24}, ensuring that proposed timetable adjustments remain within practical regulatory limits.

\subsection{Tests without hyperparameter tuning}
\label{sec:tests_without_hiperparameter_tunning}

With the first experiment, all the selected metaheuristics were tested to maximize the \gls{IM} revenue obtained based on the same 25 schedule requests. In this first test, no hyperparameter optimization was applied, selecting a standard 100-epoch optimization process and 20 individuals (for the population-based metaheuristics) and $ 100 \cdot 20 = 2000 $ epoch for the single-solution-based algorithms, i.e. \gls{SA}. The default parameters for each algorithm are shown in Table \ref{tab:params}.There are no parameters that can be adjusted (apart from epochs and population size) in the implemented \textit{MEALPY} version of \gls{CMA-ES}, \gls{GWO}, \gls{WOA} and \gls{GWO-WOA}. Thus, no values are specified in Table \ref{tab:params} for those algorithms.

\begin{table}[!ht]
\centering
\scriptsize
\caption{Default parameters in \textit{MEALPY} for each algorithm.}
\begin{adjustbox}{max width=0.45\textwidth}
\begin{tabular}{|c|l|c|}
\hline
\textbf{Algorithm} & \textbf{Parameter} & \textbf{Default Value} \\ \hline \hline
\multirow{3}{*}{\textbf{Genetic Algorithm (GA)}} & Crossover probability ($pc$) & 0.95 \\
& Mutation probability ($pm$) & 0.025 \\
& Selection method & Tournament \\ \hline
\multirow{3}{*}{\textbf{Particle Swarm Optimization (PSO)}} & Local coefficient ($c1$) & 2.05 \\
& Global coefficient ($c2$) & 2.05 \\
& Positive constant (alpha) & 0.4 \\ \hline
\multirow{2}{*}{\textbf{Simulated Annealing (SA)}} & Initial temperature & 100 \\
& Cooling rate & 0.99 \\ \hline
\multirow{2}{*}{\textbf{Differential Evolution (DE)}} & Weighting factor ($wf$) & 0.1 \\
& Crossover rate ($cr$) & 0.9 \\ \hline
\multirow{3}{*}{\makecell{\textbf{Ant Colony Optimization} \\ \textbf{Continuous (ACOR)}}} & Sample count & 25 \\
& Intensification factor & 0.5 \\
& Zeta & 1.0 \\ \hline
\makecell{\textbf{Covariance Matrix Adaptation Evolution} \\ \textbf{Strategy}} & - & - \\ \hline
\multirow{1}{*}{\textbf{Artificial Bee Colony (ABC)}} & Food source limit & 25 \\ \hline
\multirow{1}{*}{\textbf{Grey Wolf Optimizer}} & - & - \\ \hline
\multirow{1}{*}{\textbf{Whale Optimization Algorithm}} & - & - \\ \hline
\makecell{\textbf{Hybrid Grey Wolf - Whale Optimization} \\ \textbf{Algorithm}} & - & - \\ \hline
\end{tabular}
\end{adjustbox}
\label{tab:params}
\end{table}

\begin{figure*}[ht]
    \centering
    \includegraphics[width=0.8\textwidth]{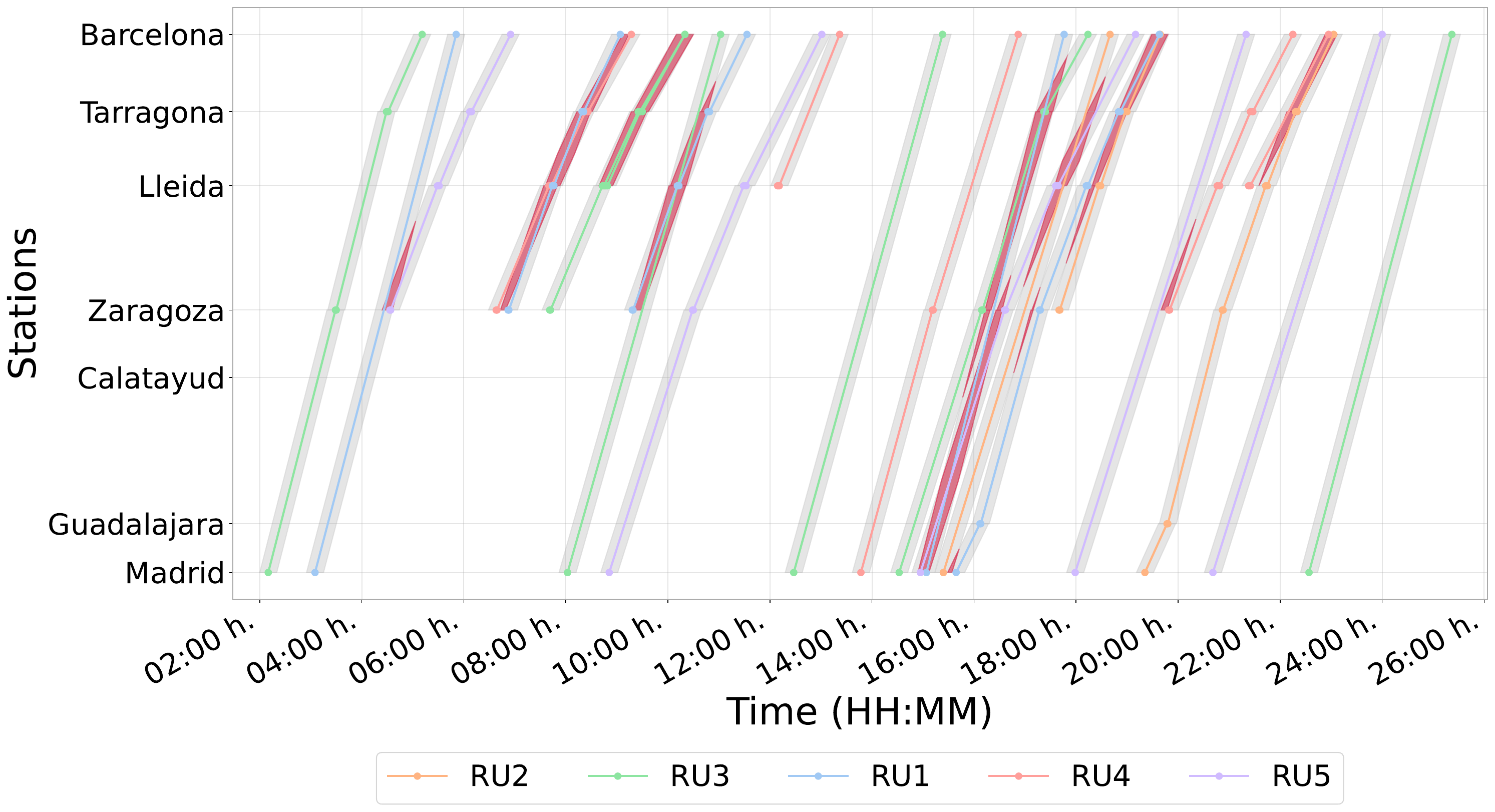}
    \caption{Marey chart with the requested schedule.}
    \label{fig:requests_25} 
\end{figure*}

First, 25 random schedule requests were generated from four different \glspl{RU}. Multiple conflicts arise in those requests. A minimum safety gap of 10 minutes was introduced between services. The requested schedules with safety gaps have been represented in Figure \ref{fig:requests_25}, where conflicts between services are highlighted in red.

\begin{figure}[ht]
  \centering
    \includegraphics[
      width=\linewidth,
    ]{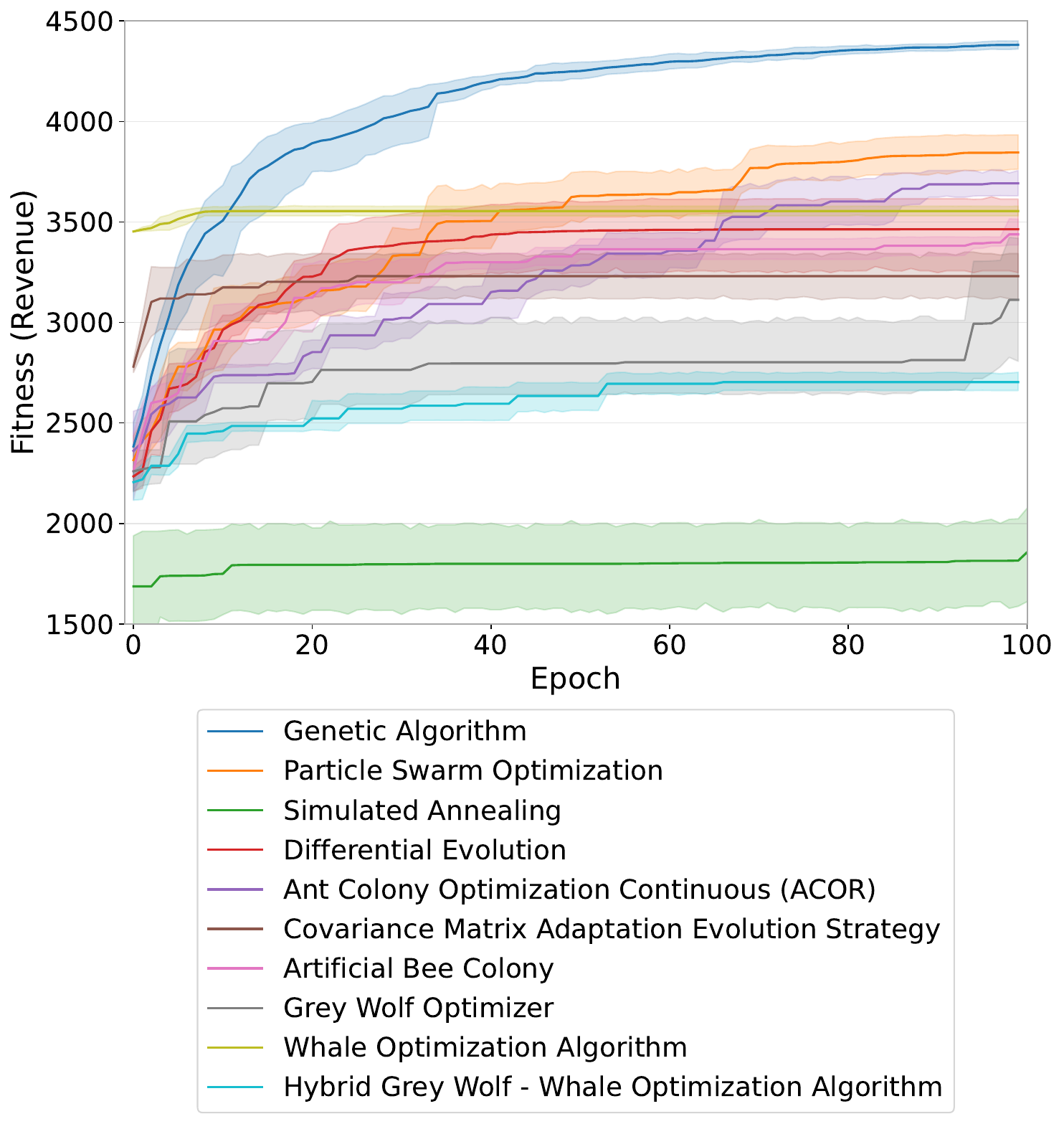}
  \caption{Convergence curves of all algorithms.}
  \label{fig:convergence}
\end{figure}

\begin{figure}[ht]
    \centering
    \includegraphics[width=0.5\textwidth]{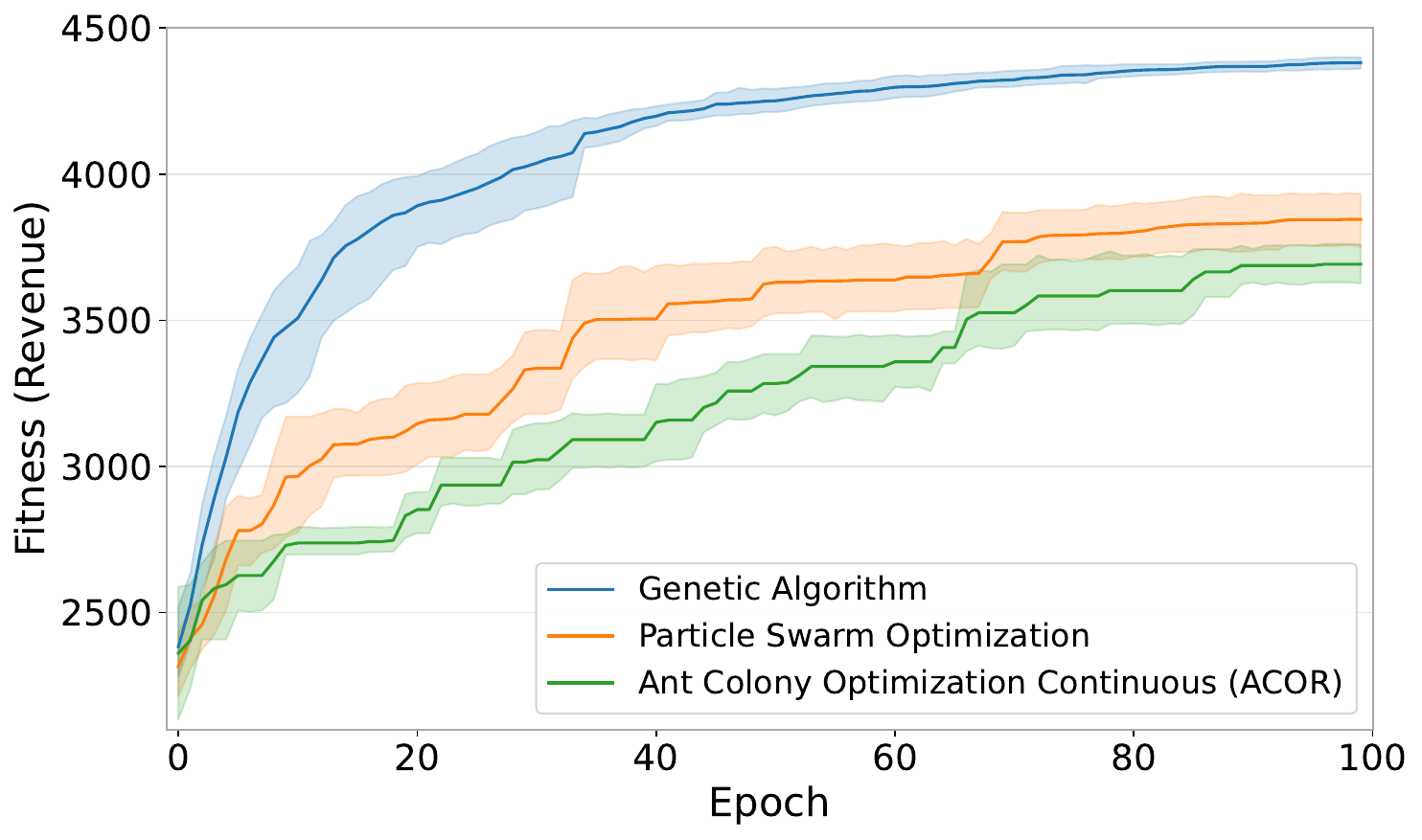}
    \caption{Convergence curves of top 3 algorithms.}
    \label{fig:convergence_top_3} 
\end{figure}

Based on this data, the different metaheuristics were applied to obtain an adjusted schedule maximizing the \gls{IM} revenue. Each algorithm is allowed to modify the departure times with a maximum difference of 10 minutes with respect to the requests, and taking into account the associated penalties (decrease in revenue) for the \gls{IM} based on the sensitivity of the \gls{RU}. The convergence analysis shown in Figure \ref{fig:convergence} revealed notable differences in the performance of the algorithms. Among the ten algorithms tested, \gls{GA}, \gls{PSO} and \gls{ACOR} demonstrated superior convergence behavior, achieving higher revenue across epochs. The \gls{GA} had the best overall performance, achieving the highest revenue with steady convergence. \gls{PSO} achieves competitive results but a slightly slower convergence rate. In turn, \gls{ACOR} gave competitive fitness values, presented more variability during its convergence. A focused comparison of the top three algorithms is shown in Figure \ref{fig:convergence_top_3}, where the smoother and faster convergence of \gls{GA} compared to \gls{PSO} and \gls{ACOR} can be observed in more detail. The remaining algorithms showed slower convergence and lower overall fitness values. These results highlight the importance of selecting optimization algorithms tailored to the characteristics of the problem, as performance varied significantly depending on the algorithm tested.

\begin{figure}[ht]
    \centering
    \includegraphics[width=0.5\textwidth]{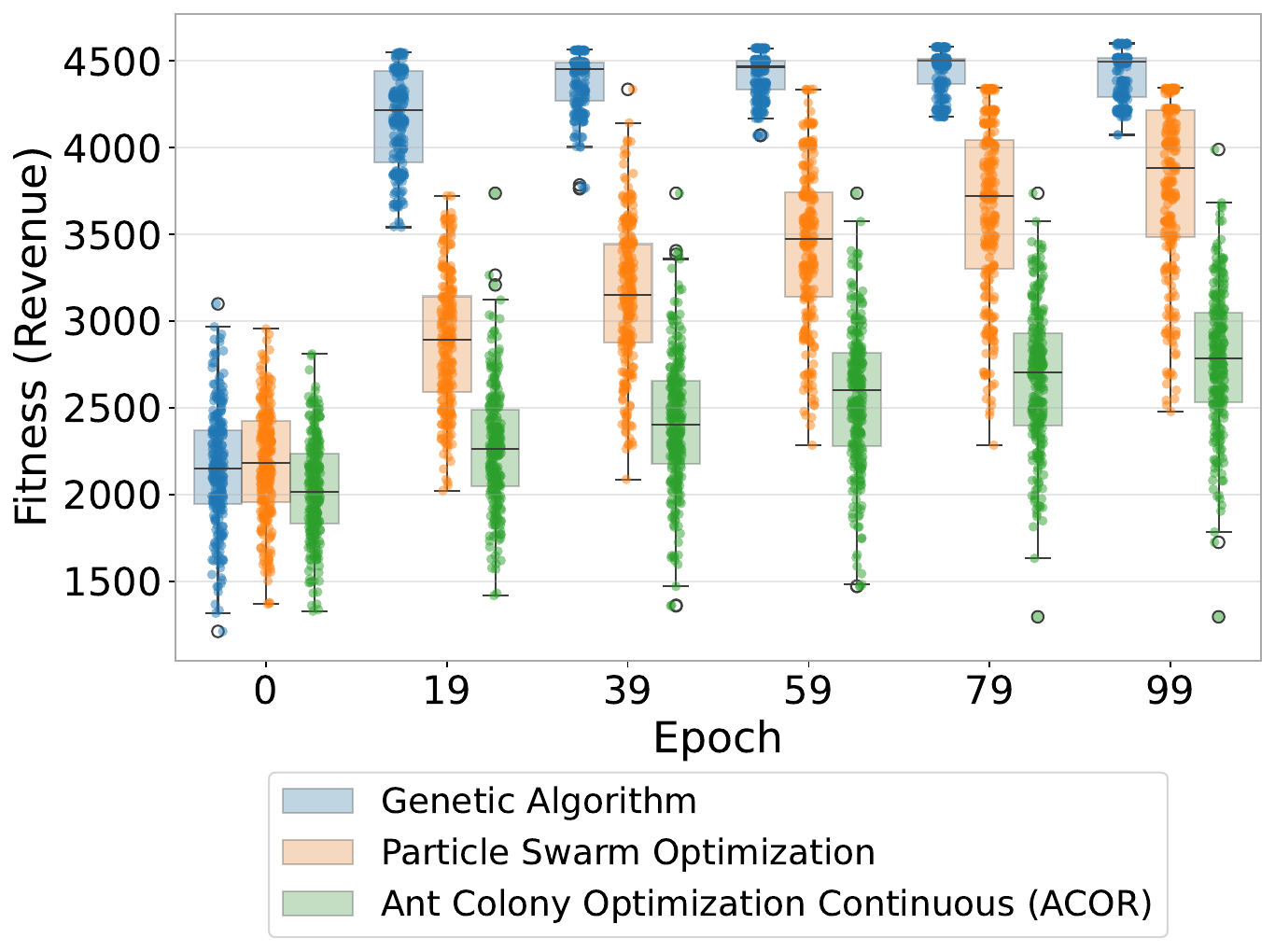}
    \caption{Scattered boxplot of population.}
    \label{fig:scattered_top_3} 
\end{figure}

The scattered boxplot analysis of the population for the top three algorithms (\gls{GA}, \gls{PSO} and \gls{ACOR}) shown in Figure \ref{fig:scattered_top_3} provides valuable insights not only into their convergence behavior but also into the evolution and dispersion of their solutions. \gls{GA} shows a significant reduction in solution dispersion as iterations progress, reflecting its ability to converge rapidly towards a narrow set of high-quality solutions. This behavior demonstrates its strong exploitation capabilities and efficient narrowing of the search space. In contrast, the dispersion of solutions in \gls{PSO} remains relatively constant throughout the iterations. This suggests that while \gls{PSO} achieves steady improvements in fitness, its population maintains diversity, which may be beneficial for prolonged exploration in dynamic or less-constrained problems. \gls{ACOR}, on the other hand, exhibits an intermediate behavior, with some reduction in dispersion over time, but not as pronounced as in \gls{GA}. This indicates a balance between exploration and exploitation, allowing \gls{ACOR} to refine solutions while maintaining a moderate level of diversity. These differences in solution dispersion underscore the unique characteristics of each algorithm, meaning that they can be complementary approaches, depending on the specific requirements of the optimization task.

Table \ref{tab:execution_data} shows the results achieved for each algorithm for each one of the five different runs, ranked by the obtained revenue (higher to lower). The results highlight the clear dominance of \gls{GA} across all metrics. It consistently achieves the highest revenue over the majority of runs, demonstrating its effectiveness in optimizing the problem at hand. However, an interesting observation arises in runs three and five of \gls{GA}, where scheduling more trains (nineteen in run five versus eighteen in run three) does not necessarily translate into higher revenue. This indicates that the quality of the solution is not solely dependent on the number of scheduled trains, but also on other factors such as minimizing schedule changes. This happens because, in order to schedule a higher number of services, it is often required to make more significant changes to the requested times, thereby incurring higher penalties. The metrics $ \Delta DT $ and $ \Delta TT $, which represent the absolute deviation in minutes of the solution's departure times and travel times, respectively, from the initially requested schedules, provide additional insights into these matters. Smaller deviations in these metrics, as achieved by \gls{GA}, result in fewer penalties due to schedule changes, making such solutions more favorable in contexts where adherence to initial schedules is critical. However, it is worth noting that larger deviations can sometimes allow for the planning of a greater number of trains, which may lead to a higher revenue for the \gls{IM}. This trade-off highlights the importance of balancing revenue maximization with adherence to a schedule. In this case, \gls{GA} excels at minimizing deviations and maximizing revenue compared with the other methods. 

\begin{table*}[htbp]
\centering
\caption{Results by run for each of the applied metaheuristics.}
\begin{adjustbox}{max width=\textwidth}
\begin{tabular}{|c|l|c|c|c|c|c|c|}
\hline
\textbf{Rank} & \textbf{Algorithm} & \textbf{Run} & \makecell{\textbf{Revenue}} & \makecell{\textbf{Execution} \\ \textbf{Time (s.)}} & \makecell{\textbf{Scheduled} \\ \textbf{Trains}} & \makecell{\textbf{$\Delta DT$} \\ \textbf{(min.)}} & \makecell{\textbf{$\Delta TT$} \\ \textbf{(min.)}} \\ \hline \hline
1 & Genetic Algorithm & 2 & \textbf{4407.02} & 25.41 & 18 & 20.00 & 20.00 \\ \hline
2 & Genetic Algorithm & 3 & 4397.00 & 25.26 & 18 & 22.00 & 14.00 \\ \hline
3 & Genetic Algorithm & 5 & 4394.21 & 26.33 & \textbf{19} & 34.00 & 37.00 \\ \hline
4 & Genetic Algorithm & 4 & 4359.11 & 25.01 & 18 & 27.00 & 26.00 \\ \hline
5 & Genetic Algorithm & 1 & 4348.73 & 24.98 & 18 & 22.00 & 33.00 \\ \hline
6 & Particle Swarm Optimization & 1 & 3972.60 & 25.94 & 17 & 48.39 & 104.10 \\ \hline
7 & Particle Swarm Optimization & 3 & 3951.24 & 25.11 & 17 & 36.46 & 99.85 \\ \hline
8 & Ant Colony Optimization Continuous (ACOR) & 2 & 3812.21 & 15.06 & 18 & 98.18 & 104.90 \\ \hline
9 & Particle Swarm Optimization & 2 & 3806.60 & 25.11 & 17 & 54.25 & 97.19 \\ \hline
10 & Particle Swarm Optimization & 5 & 3797.09 & 24.81 & 17 & 44.57 & 112.32 \\ \hline
11 & Particle Swarm Optimization & 4 & 3699.86 & 24.72 & 17 & 45.45 & 96.20 \\ \hline
12 & Ant Colony Optimization Continuous (ACOR) & 4 & 3697.80 & \textbf{15.05} & 18 & 122.62 & 105.75 \\ \hline
13 & Ant Colony Optimization Continuous (ACOR) & 5 & 3692.79 & 15.24 & 18 & 118.37 & 97.90 \\ \hline
14 & Ant Colony Optimization Continuous (ACOR) & 1 & 3670.42 & 15.08 & 18 & 104.03 & 119.96 \\ \hline
15 & Differential Evolution & 3 & 3656.84 & 25.25 & 15 & 34.58 & 104.83 \\ \hline
16 & Differential Evolution & 5 & 3609.12 & 25.13 & 18 & 52.59 & 117.38 \\ \hline
17 & Whale Optimization Algorithm & 2 & 3592.66 & 24.73 & 15 & 123.76 & 16.63 \\ \hline
18 & Ant Colony Optimization Continuous (ACOR) & 3 & 3586.21 & 15.10 & 17 & 113.61 & 88.90 \\ \hline
19 & Artificial Bee Colony & 4 & 3579.31 & 49.69 & 16 & 77.74 & 98.18 \\ \hline
20 & Whale Optimization Algorithm & 3 & 3573.69 & 25.09 & 15 & 132.91 & 12.72 \\ \hline
21 & Whale Optimization Algorithm & 4 & 3557.95 & 25.44 & 15 & 130.26 & 12.49 \\ \hline
22 & Differential Evolution & 4 & 3546.96 & 24.97 & 17 & 52.29 & 88.76 \\ \hline
23 & Grey Wolf Optimizer & 3 & 3544.65 & 24.73 & 16 & 125.16 & 105.42 \\ \hline
24 & Grey Wolf Optimizer & 5 & 3535.08 & 24.82 & 16 & 94.35 & 68.25 \\ \hline
25 & Whale Optimization Algorithm & 1 & 3523.55 & 24.77 & 15 & 145.10 & 12.52 \\ \hline
26 & Whale Optimization Algorithm & 5 & 3523.55 & 25.04 & 15 & 144.87 & 12.92 \\ \hline
27 & Artificial Bee Colony & 1 & 3451.19 & 49.93 & 18 & 110.82 & 119.77 \\ \hline
28 & Differential Evolution & 1 & 3442.21 & 25.33 & 17 & 79.68 & 100.52 \\ \hline
29 & Artificial Bee Colony & 3 & 3395.20 & 50.14 & 17 & 100.57 & 95.61 \\ \hline
30 & Artificial Bee Colony & 5 & 3387.38 & 50.40 & 18 & 117.69 & 126.26 \\ \hline
31 & Covariance Matrix Adaptation Evolution Strategy & 2 & 3385.11 & 51.76 & 15 & 36.63 & 109.43 \\ \hline
32 & Artificial Bee Colony & 2 & 3378.89 & 50.02 & 17 & 75.28 & 90.61 \\ \hline
33 & Covariance Matrix Adaptation Evolution Strategy & 3 & 3348.85 & 51.72 & 15 & 18.22 & 111.26 \\ \hline
34 & Covariance Matrix Adaptation Evolution Strategy & 5 & 3246.36 & 51.82 & 15 & 35.02 & 108.31 \\ \hline
35 & Covariance Matrix Adaptation Evolution Strategy & 1 & 3097.36 & 55.24 & 15 & 26.14 & 99.94 \\ \hline
36 & Covariance Matrix Adaptation Evolution Strategy & 4 & 3073.36 & 50.11 & 15 & 29.16 & 88.93 \\ \hline
37 & Differential Evolution & 2 & 3062.25 & 25.01 & 16 & 80.95 & 94.17 \\ \hline
38 & Grey Wolf Optimizer & 1 & 2946.60 & 24.75 & 17 & 118.04 & 105.45 \\ \hline
39 & Grey Wolf Optimizer & 2 & 2832.62 & 24.94 & 14 & 142.48 & 72.33 \\ \hline
40 & Hybrid Grey Wolf - Whale Optimization Algorithm & 2 & 2786.66 & 24.74 & 16 & 158.47 & 94.35 \\ \hline
41 & Simulated Annealing & 5 & 2762.54 & 25.94 & 16 & 16.25 & 78.38 \\ \hline
42 & Hybrid Grey Wolf - Whale Optimization Algorithm & 5 & 2734.24 & 25.02 & 17 & 164.41 & 70.62 \\ \hline
43 & Grey Wolf Optimizer & 4 & 2704.17 & 24.73 & 17 & 161.57 & 61.50 \\ \hline
44 & Hybrid Grey Wolf - Whale Optimization Algorithm & 1 & 2682.90 & 25.33 & 17 & 176.81 & 65.03 \\ \hline
45 & Hybrid Grey Wolf - Whale Optimization Algorithm & 3 & 2663.62 & 24.95 & 15 & 145.79 & 92.50 \\ \hline
46 & Simulated Annealing & 1 & 2656.66 & 26.00 & 16 & 35.46 & 54.77 \\ \hline
47 & Hybrid Grey Wolf - Whale Optimization Algorithm & 4 & 2647.82 & 24.87 & 15 & 146.80 & 77.45 \\ \hline
48 & Simulated Annealing & 4 & 2601.35 & 25.67 & 15 & 43.97 & 84.07 \\ \hline
49 & Simulated Annealing & 3 & 2507.06 & 25.95 & 15 & 33.06 & 71.05 \\ \hline
50 & Simulated Annealing & 2 & 1796.63 & 26.43 & 15 & 76.45 & 85.58 \\ \hline
\end{tabular}
\end{adjustbox}
\label{tab:execution_data}
\end{table*}

Table \ref{tab:summary_results_compact} summarizes and highlights significant differences in the performance of the algorithms, particularly in terms of revenue (\(\mu_{revenue}\)) and the number of scheduled trains (\(\mu_{trains}\)). \gls{GA} achieved the highest mean revenue (\(4381.21\)), significantly outperforming all other algorithms with a consistent standard deviation (\(\sigma_{revenue} = 25.63\)) and no variability in the number of scheduled trains (\(\mu_{trains} = 18\), \(\sigma_{trains} = 0\)). Moreover, \gls{PSO} and \gls{ACOR} are in second and third place, their mean revenues (\(3845.48\) and \(3691.89\), respectively) differ significantly from \gls{GA}. Notably, \gls{ACOR} matches \gls{GA} in the number of scheduled trains (\(18\)) but achieves lower revenues, due to less efficient scheduling and higher deviations as seen in Table \ref{tab:execution_data}. Further down the rankings, \gls{WOA} and \gls{DE} show substantial revenue reduction (\(3554.28\) and \(3463.48\), respectively) and fewer scheduled trains. The lowest-ranked algorithms, such as \gls{SA} and \gls{GWO-WOA}, exhibit not only lower revenues but also greater variability in their results, as indicated by their larger standard deviations.

\begin{table*}[!ht]
\centering
\caption{Summary of results.}
\begin{adjustbox}{max width=\textwidth}
\begin{tabular}{|c|l|r|r|r|r|r|r|}
\hline
\textbf{Rank} & \textbf{Algorithm} & \(\mu_{time}\) & \(\sigma_{time}\) & \(\mu_{revenue}\) & \(\sigma_{revenue}\) & \(\mu_{trains}\) & \(\sigma_{trains}\) \\ \hline \hline
1 & Genetic Algorithm & 25.40 & 0.55 & \textbf{4381.21} & \textbf{25.63} & 18 & 0 \\ \hline
2 & Particle Swarm Optimization & 25.14 & 0.48 & 3845.48 & 114.46 & 17 & 0 \\ \hline
3 & Ant Colony Optimization (ACOR) & 15.11 & 0.08 & 3691.89 & 80.84 & 18 & 0 \\ \hline
4 & Whale Optimization Algorithm & 25.01 & 0.29 & 3554.28 & 30.63 & 15 & 0 \\ \hline
5 & Differential Evolution & 25.14 & 0.15 & 3463.48 & 238.22 & 17 & 1 \\ \hline
6 & Artificial Bee Colony & 50.03 & 0.26 & 3438.39 & 83.71 & 17 & 1 \\ \hline
7 & Covariance Matrix Adaptation & 52.13 & 1.88 & 3230.21 & 141.93 & 15 & 0 \\ \hline
8 & Grey Wolf Optimizer & 24.79 & 0.09 & 3112.63 & 399.35 & 16 & 1 \\ \hline
9 & Hybrid Grey Wolf - Whale & 24.98 & 0.22 & 2703.05 & 56.95 & 16 & 1 \\ \hline
10 & Simulated Annealing & 26.00 & 0.27 & 2464.85 & 384.82 & 15 & 1 \\ \hline
\end{tabular}
\end{adjustbox}
\label{tab:summary_results_compact}
\end{table*}

\begin{figure}[ht]
    \centering
    \includegraphics[width=\linewidth]{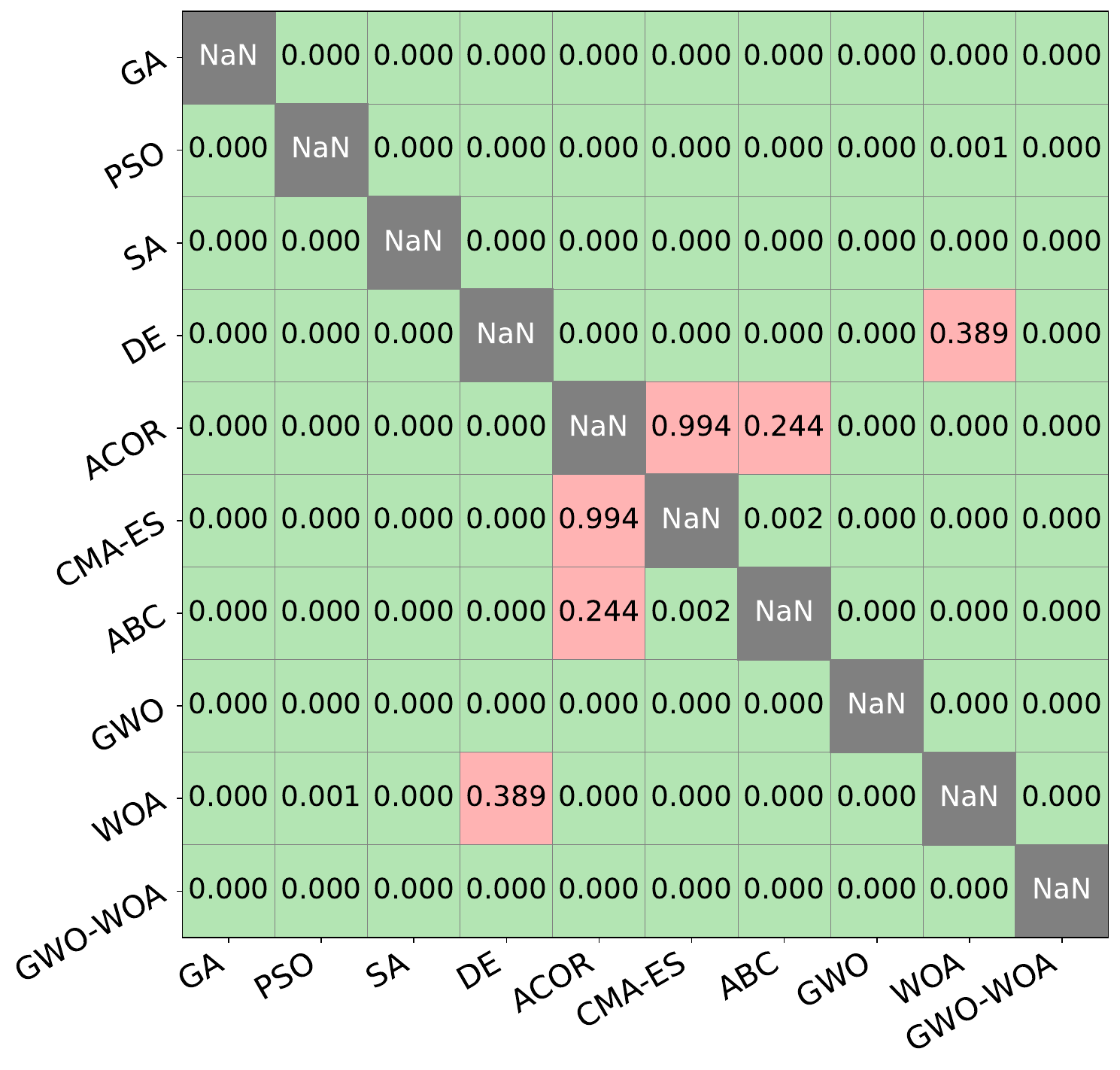}
    \caption{Matrix of p-values obtained from the Wilcoxon test.}
    \label{fig:p_values} 
\end{figure}

Figure \ref{fig:p_values} shows the statistical comparison based on $p$-values using the Wilcoxon test \cite{Derrac12,Garcia10} for paired samples. This test aims to determine whether there are significant differences in the performance between two related samples, making it well suited for comparing algorithm results across paired runs. Values below 0.05 are displayed in green, indicating a significant difference in favor of one algorithm, while values above 0.05 are shown in red, indicating no significant difference. An algorithm is not compared with itself. Therefore, the main diagonal of the matrix has empty values. The results reveal significant differences between most pairs of algorithms at a significance level of 95\%. However, certain pairs, such as \gls{ACOR} versus \gls{CMA-ES} and \gls{ABC} exhibit $p$-values greater than 0.05 ($ p=0.994 $ and $ p=0.244  $ respectively), indicating no statistically significant difference in their results. Similarly, \gls{DE} versus \gls{WOA} yields a $p$-values of $ p=0.389 $, suggesting comparable performance. These non-significant differences may be due to overlapping performance in certain problem instances, where both methods reach similar solutions.

\begin{figure*}[ht]
    \centering
    \includegraphics[width=0.8\textwidth]{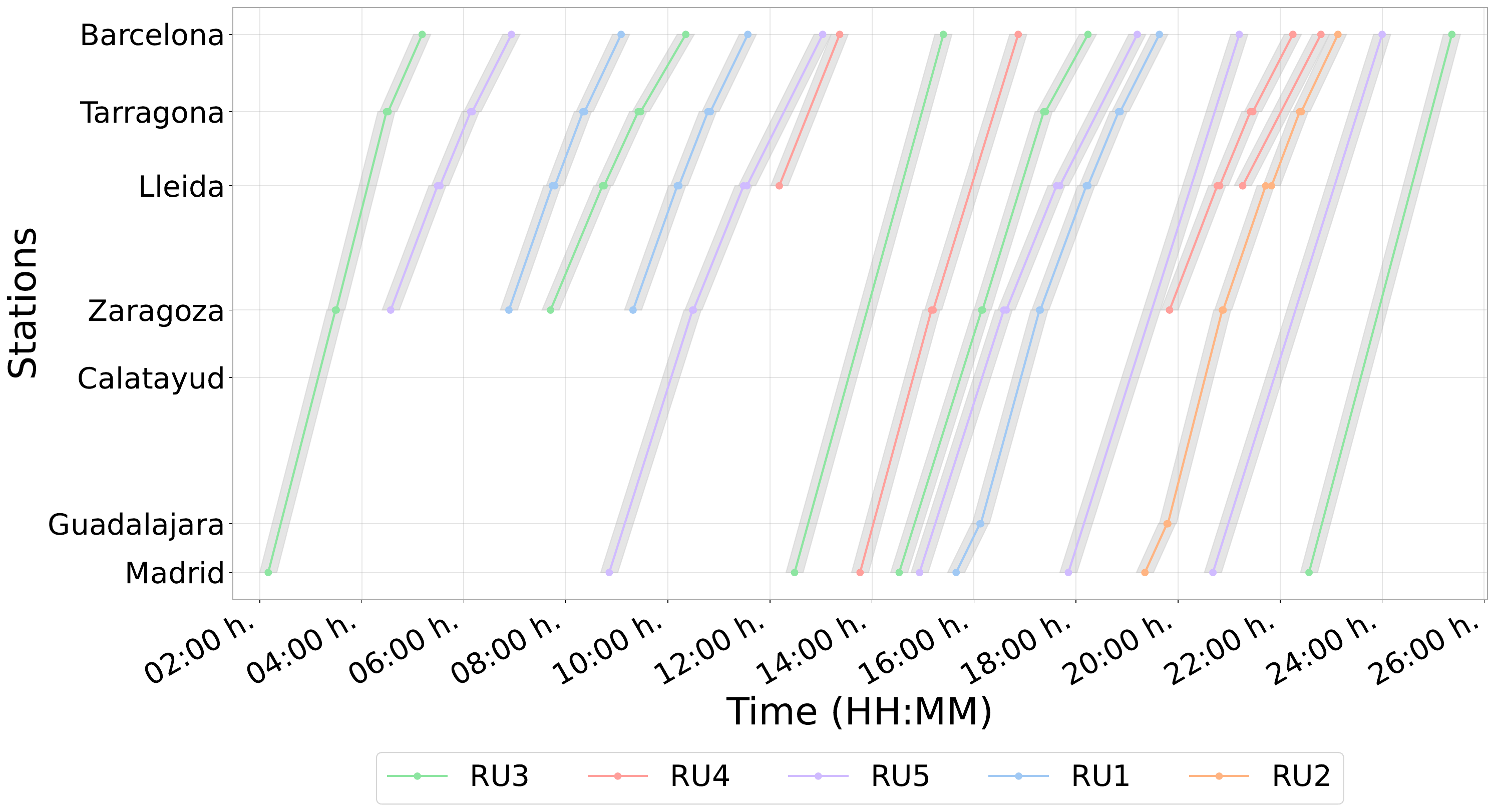}
    \caption{Marey chart with the solution obtained by the \gls{GA} in the second run.}
    \label{fig:ga_marey_Without} 
\end{figure*}

The best solution achieved with the \gls{GA} in the second run can be seen in Figure \ref{fig:ga_marey_Without}. It can be observed how the algorithm manages to balance the compromise between scheduling more services and minimizing changes to the original schedule requests, making the minimum adjustment required to meet safety constraints. It can also be seen how some services could not be scheduled based on the restrictions established in the problem.

\subsection{Tests with hyperparameter tuning}
\label{sec:tests_witht_hiperparameter_tunning}

In this second test, the two hyperparameters common to all the algorithms used (except \gls{SA}), the number of iterations and the number of individuals, were adjusted. This was accomplished using the HyperOpt Python package \cite{Bergstra13,Hutter19}. The results from Table \ref{tab:hyperparameter_results} illustrate the tuned hyperparameters for each algorithm. These adjusted values were determined based on the exploration range of 100 to 500 epochs and 10 to 100 individuals, with the exception of \gls{SA}, which is a single-solution algorithm. Instead, the number of iterations for \gls{SA} was independently set to 45000 to ensure comparable computational effort. \gls{GA} and \gls{PSO} required the maximum number of epochs (500), suggesting their effectiveness relies on extended iterations to achieve optimal results. However, their individual population sizes (70 for \gls{GA} and 80 for \gls{PSO}) reflect a balance between population diversity and computational feasibility. Interestingly, the \gls{ACOR} algorithm uses the smallest population size (10 individuals) but requires the maximum number of epochs (500). This reflects its unique mechanism, where fewer individuals suffice to produce competitive results over prolonged iterations. On the other hand, \gls{DE} and \gls{GWO} employ the maximum population size (100 individuals), but with fewer epochs in the case of \gls{DE} (300).

\begin{table}[!ht]
\centering
\caption{Hyperparameter tuning results.}
\begin{adjustbox}{max width=0.45\textwidth}
\begin{tabular}{|c|c|c|}
\hline
\textbf{Algorithm} & \textbf{Epochs} & \textbf{Individuals} \\ \hline \hline
Genetic Algorithm & 500 & 70 \\ \hline
Particle Swarm Optimization & 500 & 80 \\ \hline
Simulated Annealing & 45000 & 1 \\ \hline
Differential Evolution & 300 & 100 \\ \hline
\makecell{Ant Colony Optimization Continuous \\ (ACOR)} & 500 & 10 \\ \hline
\makecell{Covariance Matrix Adaptation Evolution \\ Strategy} & 250 & 80 \\ \hline
Artificial Bee Colony & 450 & 60 \\ \hline
Grey Wolf Optimizer & 500 & 100 \\ \hline
Whale Optimization Algorithm & 400 & 80 \\ \hline
\makecell{Hybrid Grey Wolf - Whale Optimization \\ Algorithm} & 450 & 50 \\ \hline
\end{tabular}
\end{adjustbox}
\label{tab:hyperparameter_results}
\end{table}

\begin{figure}[ht]
  \centering
    \includegraphics[
      width=\linewidth
    ]{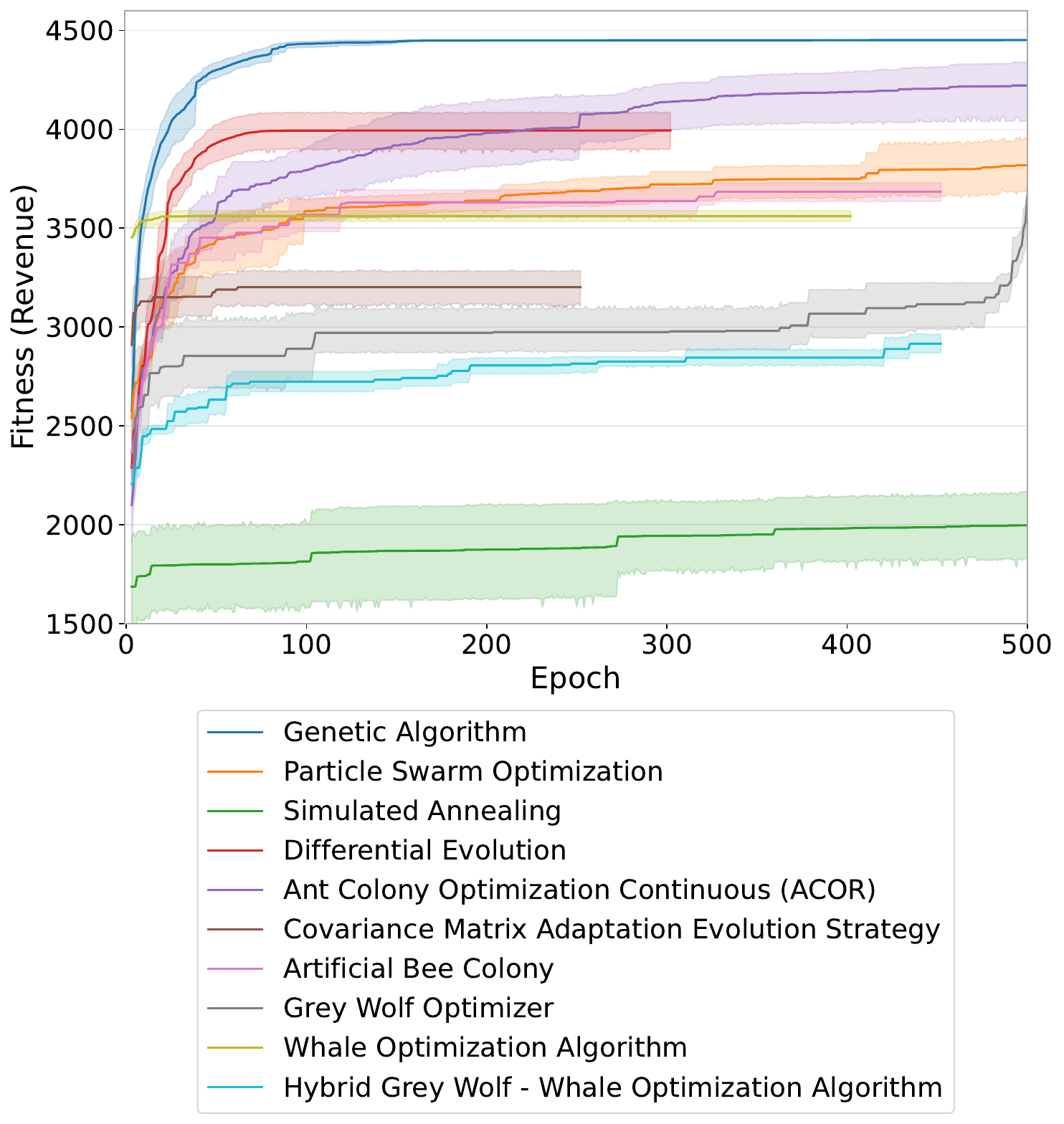}
  \caption{Convergence curves of all algorithms with hyperparameter tuning.}
  \label{fig:convergence_opti}
\end{figure}

\begin{figure}[ht]
    \centering
    \includegraphics[width=0.5\textwidth]{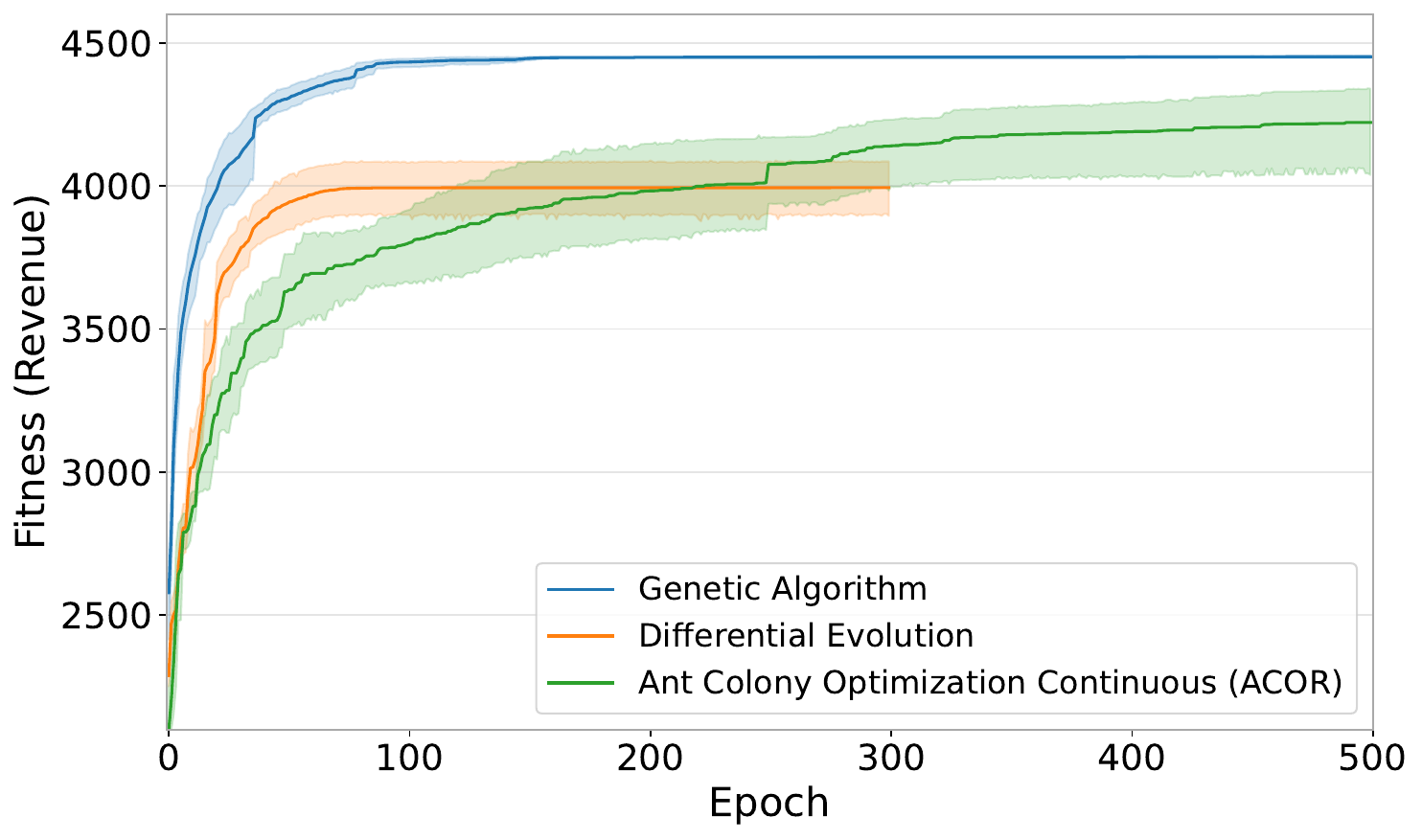 }
    \caption{Convergence curves of top 3 algorithms with hyperparameter tuning.}
    \label{fig:convergence_opti_top3} 
\end{figure}

The convergence curves for the second experiment, shown in Figures \ref{fig:convergence_opti} and \ref{fig:convergence_opti_top3}, reveal significant differences in algorithm performance after hyperparameter tuning compared to the first experiment without tuning. Figure \ref{fig:convergence_opti} displays the convergence behavior of all 10 algorithms. \gls{GA} maintains its position as the best-performing algorithm, achieving the highest fitness values across all epochs. However, its convergence is faster and more stable compared to the untuned experiment, demonstrating the impact of optimized hyperparameters. \gls{ACOR} and \gls{DE} also show notable improvements, reaching higher fitness values with reduced variability. Figure \ref{fig:convergence_opti_top3} focuses on the top three algorithms (\gls{GA}, \gls{ACOR}, and \gls{DE}) with tuned hyperparameters. \gls{GA} continues to dominate with rapid convergence and stabilization at the highest revenue levels. \gls{ACOR} and \gls{DE} exhibit competitive performance, with \gls{DE} converging faster in earlier epochs and \gls{ACOR} displaying a steady improvement across all iterations. These results contrast with the first experiment, where \gls{DE} lagged behind \gls{PSO} in overall performance. The improvements observed in \gls{DE} further reinforce the benefits of tailored hyperparameters, and also remark that even high-performing algorithms like \gls{GA} can benefit from careful hyperparameter adjustment to further optimize their performance.

\begin{figure}[ht]
    \centering
    \includegraphics[width=0.5\textwidth]{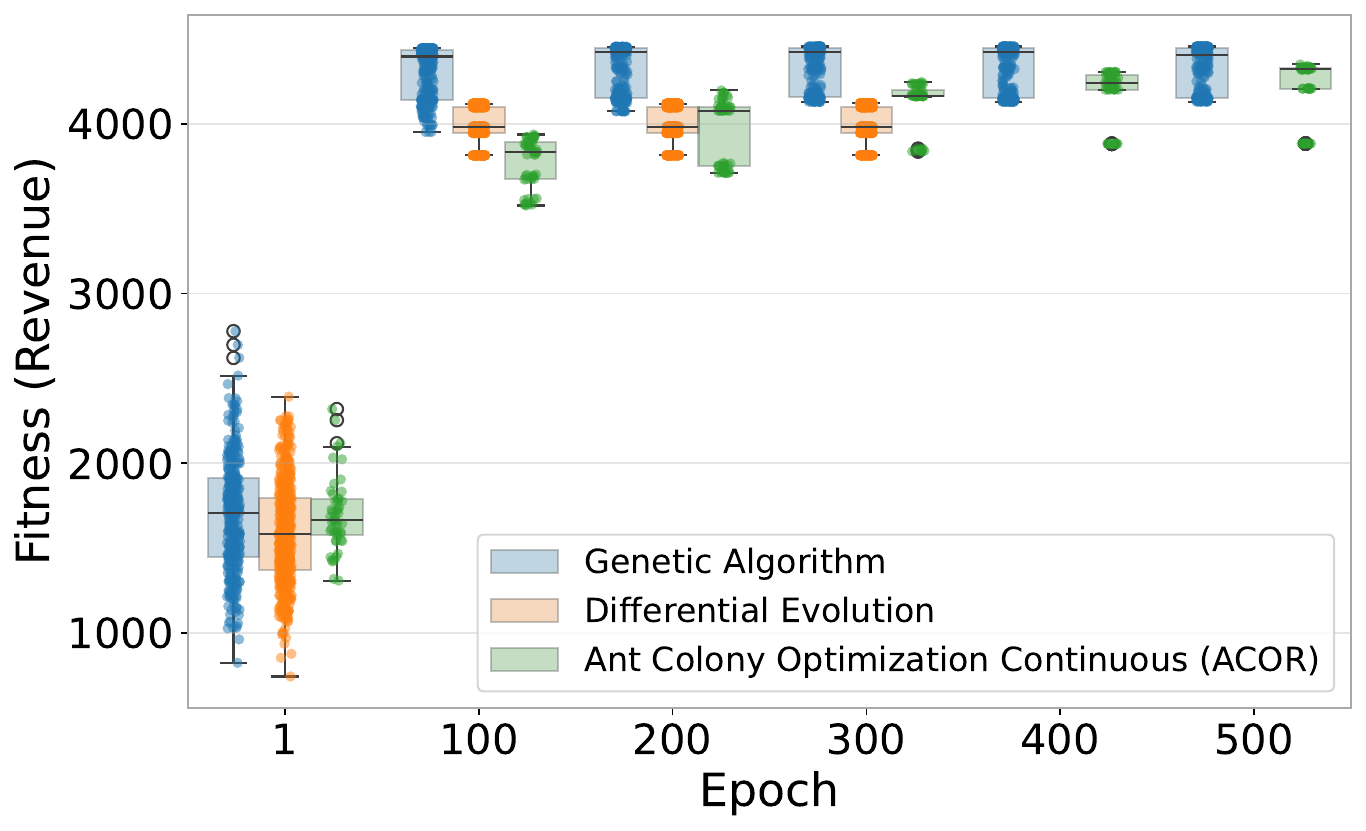}
    \caption{Scattered boxplot of population for top 3 algorithms with hyperparameter tuning.}
    \label{fig:scattered_boxplot_opti_top_3} 
\end{figure}

The scattered boxplot analysis for the top three algorithms (\gls{GA}, \gls{DE}, and \gls{ACOR}) after hyperparameter tuning, shown in Figure \ref{fig:scattered_boxplot_opti_top_3}, provides further insights into their population dynamics and convergence behavior. The \gls{GA} stands out as the algorithm with the most rapid and stable convergence, as indicated by the narrowing dispersion of solutions over the epochs. Its population quickly converges to a narrow range of high-revenue solutions, demonstrating its strong exploitation capabilities. This behavior aligns with the results observed in the convergence curves, confirming \gls{GA}’s dominance in both speed and quality of convergence. \gls{DE} exhibits a less gradual reduction in solution dispersion. Its population stays practically unchanged for the last 200 epochs of the optimization process. However, the overall improvement highlights the benefits of hyperparameter tuning, as \gls{DE} performed less effectively in the first experiment. \gls{ACOR}, while showing reduced dispersion in the first stages compared to \gls{GA} and \gls{DE}, demonstrates sustained improvement across all epochs. The comparison with the first experiment reveals that hyperparameter tuning significantly enhances population dynamics for all three algorithms. \gls{GA} achieves faster convergence with reduced dispersion, while \gls{DE} and \gls{ACOR} display improved performance and more consistent convergence trends. These findings underscore the critical role of hyperparameter adjustment in optimizing algorithmic performance.

The detailed results in Table \ref{tab:execution_data_opti} reveal important observations regarding the performance of the algorithms after hyperparameter tuning. The Genetic Algorithm (\gls{GA}) consistently dominates the rankings, achieving the highest revenue (\textbf{4457.33}) in run four while scheduling the highest number of trains (\textbf{19}). This confirms \gls{GA} as the best-performing algorithm in terms of revenue maximization and operational efficiency. \gls{ACOR} ranks second overall, achieving a revenue of 4351.49 in run four, with competitive performance across all metrics. However, its higher deviations in \(\Delta DT\) and \(\Delta TT\) indicate less precision in adhering to requested schedules. This may limit its potential for achieving the highest revenues compared to \gls{GA}. Differential Evolution (\gls{DE}) and Particle Swarm Optimization (\gls{PSO}) return mixed results, with \gls{DE} achieving higher revenues (up to 4122.87 in run one) but requiring longer run times compared to \gls{PSO}. The remaining algorithms, such as \gls{GWO} and \gls{ABC}, perform less competitively, with lower revenues and higher variability in both run time and schedule deviations.

\begin{table*}[htbp]
\centering
\caption{Results by run for each of the applied metaheuristics with hyperparameter tunning.}
\begin{adjustbox}{max width=\textwidth}
\begin{tabular}{|c|l|c|c|c|c|c|c|}
\hline
\textbf{Rank} & \textbf{Algorithm} & \textbf{Run} & \makecell{\textbf{Revenue}} & \makecell{\textbf{Execution} \\ \textbf{Time (s.)}} & \makecell{\textbf{Scheduled} \\ \textbf{Trains}} & \makecell{\textbf{$\Delta DT$} \\ \textbf{(min.)}} & \makecell{\textbf{$\Delta TT$} \\ \textbf{(min.)}} \\ \hline \hline
1 & Genetic Algorithm & 4 & \textbf{4457.33} & 174.37 & \textbf{19} & 27.00 & 19.00 \\ \hline
2 & Genetic Algorithm & 1 & 4455.60 & 175.23 & 18 & 18.00 & 7.00 \\ \hline
3 & Genetic Algorithm & 3 & 4448.94 & 174.59 & 18 & 18.00 & 11.00 \\ \hline
4 & Genetic Algorithm & 5 & 4448.12 & 172.57 & 18 & 17.00 & 9.00 \\ \hline
5 & Genetic Algorithm & 2 & 4448.12 & 175.70 & 18 & 17.00 & 9.00 \\ \hline
6 & Ant Colony Optimization Continuous (ACOR) & 4 & 4351.49 & 75.53 & 18 & 54.00 & 44.00 \\ \hline
7 & Ant Colony Optimization Continuous (ACOR) & 3 & 4335.49 & 75.96 & 17 & 54.00 & 27.00 \\ \hline
8 & Ant Colony Optimization Continuous (ACOR) & 5 & 4327.14 & 75.40 & 18 & 29.00 & 39.00 \\ \hline
9 & Ant Colony Optimization Continuous (ACOR) & 1 & 4211.83 & 75.56 & 18 & 46.00 & 67.00 \\ \hline
10 & Differential Evolution & 1 & 4122.87 & 153.95 & 17 & 34.00 & 62.00 \\ \hline
11 & Differential Evolution & 5 & 4099.90 & 150.23 & 18 & 46.00 & 66.00 \\ \hline
12 & Particle Swarm Optimization & 3 & 4053.13 & 198.95 & 16 & 31.00 & 54.00 \\ \hline
13 & Differential Evolution & 2 & 3984.19 & 149.62 & 16 & 22.00 & 38.00 \\ \hline
14 & Differential Evolution & 3 & 3948.05 & 148.17 & 18 & 74.00 & 102.00 \\ \hline
15 & Ant Colony Optimization Continuous (ACOR) & 2 & 3884.89 & 74.56 & 19 & 86.00 & 48.00 \\ \hline
16 & Grey Wolf Optimizer & 1 & 3871.34 & 243.73 & 18 & 103.00 & 61.00 \\ \hline
17 & Particle Swarm Optimization & 1 & 3869.11 & 194.38 & 18 & 75.00 & 90.00 \\ \hline
18 & Differential Evolution & 4 & 3815.88 & 150.04 & 17 & 57.00 & 85.00 \\ \hline
19 & Particle Swarm Optimization & 4 & 3795.09 & 200.36 & 17 & 31.00 & 47.00 \\ \hline
20 & Particle Swarm Optimization & 5 & 3783.78 & 200.56 & 16 & 41.00 & 42.00 \\ \hline
21 & Grey Wolf Optimizer & 2 & 3761.66 & 246.57 & 17 & 91.00 & 70.00 \\ \hline
22 & Artificial Bee Colony & 4 & 3737.54 & 269.38 & 18 & 90.00 & 111.00 \\ \hline
23 & Artificial Bee Colony & 2 & 3723.30 & 269.93 & 18 & 90.00 & 108.00 \\ \hline
24 & Artificial Bee Colony & 1 & 3722.66 & 267.53 & 16 & 63.00 & 86.00 \\ \hline
25 & Grey Wolf Optimizer & 4 & 3683.14 & 243.73 & 17 & 85.00 & 78.00 \\ \hline
26 & Artificial Bee Colony & 5 & 3636.26 & 272.58 & 17 & 81.00 & 91.00 \\ \hline
27 & Grey Wolf Optimizer & 3 & 3632.77 & 242.28 & 15 & 108.00 & 53.00 \\ \hline
28 & Grey Wolf Optimizer & 5 & 3617.46 & 246.01 & 16 & 87.00 & 53.00 \\ \hline
29 & Whale Optimization Algorithm & 3 & 3605.23 & 156.73 & 15 & 116.00 & 2.00 \\ \hline
30 & Artificial Bee Colony & 3 & 3603.95 & 271.26 & 17 & 89.00 & 109.00 \\ \hline
31 & Particle Swarm Optimization & 2 & 3593.19 & 196.87 & 15 & 27.00 & 79.00 \\ \hline
32 & Whale Optimization Algorithm & 1 & 3575.73 & 155.46 & 15 & 138.00 & 2.00 \\ \hline
33 & Whale Optimization Algorithm & 2 & 3575.59 & 155.27 & 15 & 138.00 & 2.00 \\ \hline
34 & Whale Optimization Algorithm & 4 & 3525.50 & 155.61 & 15 & 144.00 & 4.00 \\ \hline
35 & Whale Optimization Algorithm & 5 & 3523.55 & 155.56 & 15 & 154.00 & 4.00 \\ \hline
36 & Covariance Matrix Adaptation Evolution Strategy & 5 & 3319.07 & 208.35 & 16 & 32.00 & 90.00 \\ \hline
37 & Simulated Annealing & 1 & 3313.75 & 224.26 & 16 & 9.00 & 28.00 \\ \hline
38 & Covariance Matrix Adaptation Evolution Strategy & 1 & 3268.95 & 200.87 & 15 & 23.00 & 104.00 \\ \hline
39 & Simulated Annealing & 5 & 3253.25 & 225.14 & 16 & 24.00 & 56.00 \\ \hline
40 & Covariance Matrix Adaptation Evolution Strategy & 3 & 3245.69 & 213.87 & 16 & 31.00 & 100.00 \\ \hline
41 & Simulated Annealing & 4 & 3223.27 & 224.05 & 16 & 54.00 & 59.00 \\ \hline
42 & Simulated Annealing & 3 & 3148.82 & 225.09 & 16 & 27.00 & 47.00 \\ \hline
43 & Covariance Matrix Adaptation Evolution Strategy & 4 & 3127.29 & 203.37 & 14 & 40.00 & 98.00 \\ \hline
44 & Covariance Matrix Adaptation Evolution Strategy & 2 & 3046.82 & 229.10 & 16 & 34.00 & 93.00 \\ \hline
45 & Hybrid Grey Wolf - Whale Optimization Algorithm & 1 & 2995.01 & 109.81 & 15 & 162.00 & 62.00 \\ \hline
46 & Hybrid Grey Wolf - Whale Optimization Algorithm & 4 & 2953.96 & 110.50 & 16 & 157.00 & 79.00 \\ \hline
47 & Hybrid Grey Wolf - Whale Optimization Algorithm & 3 & 2917.10 & 108.88 & 16 & 152.00 & 66.00 \\ \hline
48 & Hybrid Grey Wolf - Whale Optimization Algorithm & 2 & 2860.35 & 109.71 & 16 & 172.00 & 81.00 \\ \hline
49 & Hybrid Grey Wolf - Whale Optimization Algorithm & 5 & 2853.85 & 110.14 & 16 & 172.00 & 103.00 \\ \hline
50 & Simulated Annealing & 2 & 2624.79 & 220.20 & 16 & 80.00 & 105.00 \\ \hline
\end{tabular}
\end{adjustbox}
\label{tab:execution_data_opti}
\end{table*}

Table \ref{tab:summary_results_compactHIPER} provides a concise overview of the performance metrics for each algorithm after hyperparameter tuning. \gls{GA} continues to outperform all other methods, both in terms of revenue and solution consistency. It remains the clear leader, achieving the highest mean revenue (\(\mu_{revenue} = 4451.62\)) with the lowest standard deviation (\(\sigma_{revenue} = 4.48\)). This consistent performance, coupled with zero variability in the number of scheduled trains (\(\sigma_{trains} = 0\)), highlights \gls{GA}’s robustness and efficiency. Ant Colony Optimization Continuous (\gls{ACOR}) ranks second, delivering competitive revenue (\(\mu_{revenue} = 4222.17\)) while maintaining a moderate standard deviation (\(\sigma_{revenue} = 196.50\)). Notably, \gls{ACOR} schedules the same average number of trains as \gls{GA} (\(\mu_{trains} = 18\)), suggesting its ability to effectively utilize available resources, though with slightly higher variability (\(\sigma_{trains} = 1\)). Differential Evolution (\gls{DE}) takes third place with a mean revenue of 3994.18 and a slightly higher standard deviation (\(\sigma_{revenue} = 124.25\)). Other algorithms, \gls{PSO} and \gls{GWO}, achieve lower mean revenues (\(3818.86\) and \(3713.27\), respectively), but their performance variability remains moderate. Algorithms ranking lower, such as \gls{CMA-ES} and \gls{GWO-WOA}, give lower mean revenues and higher variability in results. While \gls{CMA-ES} maintains relatively stable run times, its inability to match the top-performing algorithms in revenue highlights its limitations in this optimization scenario.

\begin{table*}[htbp]
\centering
\caption{Summary of results with hyperparemeter tunning.}
\begin{adjustbox}{max width=\textwidth}
\begin{tabular}{|c|l|c|c|c|c|c|c|}
\hline
\textbf{Rank} & \textbf{Algorithm} & \(\mu_{time}\) & \(\sigma_{time}\) & \(\mu_{revenue}\) & \(\sigma_{revenue}\) & \(\mu_{trains}\) & \(\sigma_{trains}\) \\ \hline \hline
1 & Genetic Algorithm & 174.49 & 1.20 & \textbf{4451.62} & \textbf{4.48} & 18 & 0 \\ \hline
2 & Ant Colony Optimization Continuous (ACOR) & 75.40 & 0.52 & 4222.17 & 196.50 & 18 & 1 \\ \hline
3 & Differential Evolution & 150.40 & 2.14 & 3994.18 & 124.25 & 17 & 1 \\ \hline
4 & Particle Swarm Optimization & 198.22 & 2.61 & 3818.86 & 165.93 & 16 & 1 \\ \hline
5 & Grey Wolf Optimizer & 244.46 & 1.78 & 3713.27 & 104.76 & 17 & 1 \\ \hline
6 & Artificial Bee Colony & 270.13 & 1.91 & 3684.74 & 60.39 & 17 & 1 \\ \hline
7 & Whale Optimization Algorithm & 155.72 & 0.58 & 3561.12 & 35.53 & 15 & 0 \\ \hline
8 & Covariance Matrix Adaptation Evolution Strategy & 211.11 & 11.22 & 3201.57 & 111.53 & 15 & 1 \\ \hline
9 & Simulated Annealing & 223.75 & 2.05 & 3112.78 & 279.18 & 16 & 0 \\ \hline
10 & Hybrid Grey Wolf - Whale Optimization Algorithm & 109.81 & 0.60 & 2916.05 & 60.51 & 16 & 0 \\ \hline
\end{tabular}
\end{adjustbox}
\label{tab:summary_results_compactHIPER}
\end{table*}

\begin{figure}[ht]
    \centering
    \includegraphics[width=0.45\textwidth]{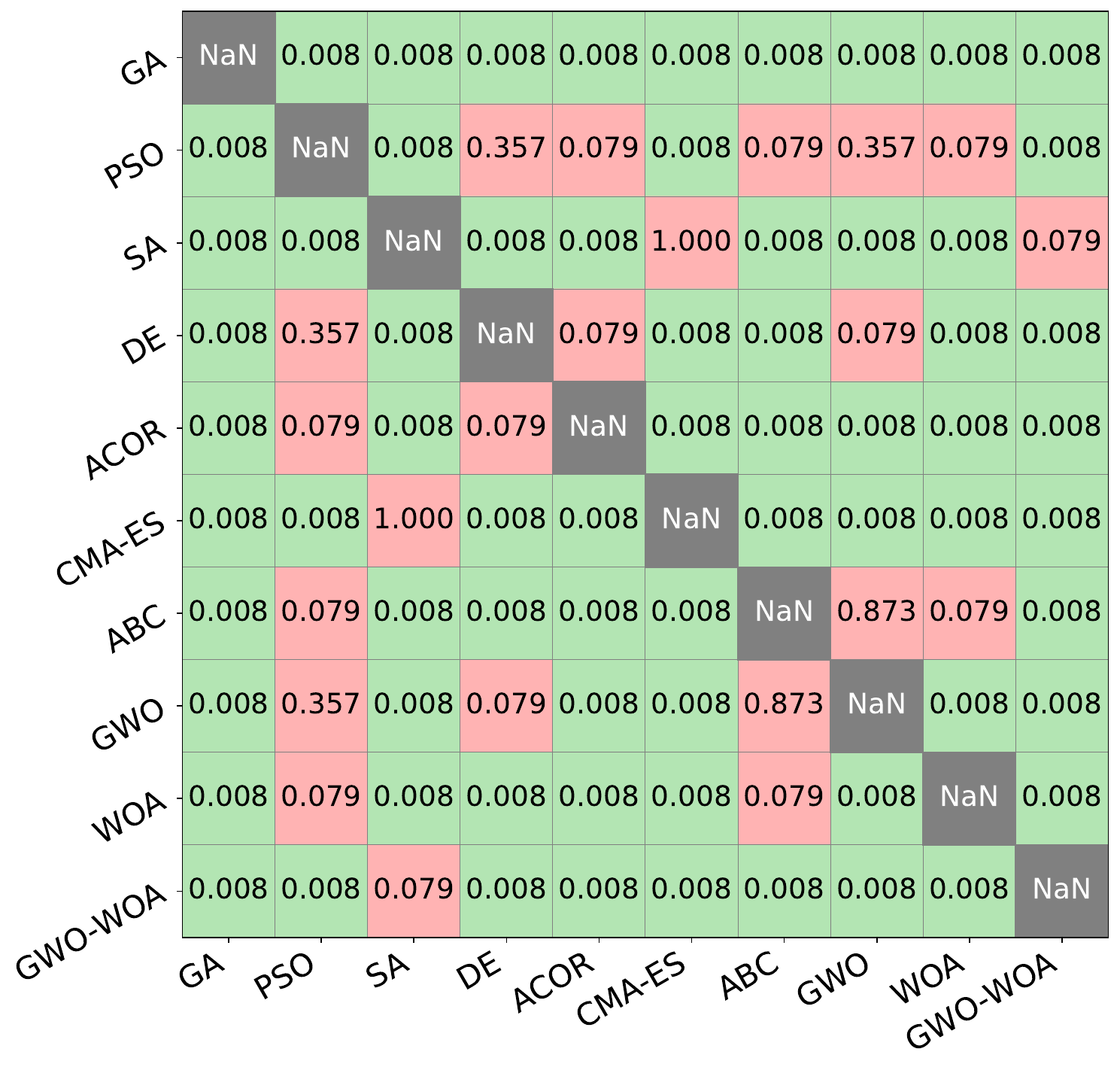}
    \caption{Matrix of $p$-values obtained from the Kolmogorov-Smirnov test.}
    \label{fig:p_values_opti} 
\end{figure}

Figure \ref{fig:p_values_opti} illustrates the statistical differences in performance between pairs of algorithms based on their best fitness values across five independent runs. The Kolmogorov-Smirnov test \cite{Kolmogorov33, Smirnov48} was used to determine whether the distributions of fitness values for each pair of algorithms differ significantly, with a significance level of 95\%. The choice of the Kolmogorov-Smirnov test for the second analysis was motivated by the nature of the comparison and the properties of the test itself. Unlike the Wilcoxon signed-rank test, the Kolmogorov-Smirnov test is more suitable for analizing differences between two independent samples, including differences in shape. As with the first experiment, values below 0.05 are displayed in green, indicating a significant difference in favor of one algorithm, while values above 0.05 are shown in red, indicating no significant difference. The results reveal that most algorithm pairs exhibit statistically significant differences ($p$-values $ < 0.05$). Notably, the \gls{GA} consistently shows significant superiority over all other algorithms. Similarly, algorithms such as \gls{DE}, \gls{ACOR}, and \gls{ABC} also show significant differences as compared to others.

\begin{table}[!ht]
\centering
\scriptsize
\caption{\parbox{0.35\textwidth}{Kolmogorov-Smirnov test p-values comparing results with and without hyperparameter tuning for each algorithm.}}
\begin{adjustbox}{max width=0.45\textwidth}
\begin{tabular}{|l|c|}
\hline
\textbf{Algorithm} & \textbf{p-value} \\ \hline \hline
Genetic Algorithm & \textbf{0.008} \\ \hline
Particle Swarm Optimization & 0.873 \\ \hline
Simulated Annealing & 0.079 \\ \hline
Differential Evolution & \textbf{0.008} \\ \hline
Ant Colony Optimization Continuous (ACOR) & \textbf{0.008} \\ \hline
Covariance Matrix Adaptation Evolution Strategy & 0.873 \\ \hline
Artificial Bee Colony & \textbf{0.008} \\ \hline
Grey Wolf Optimizer & \textbf{0.008} \\ \hline
Whale Optimization Algorithm & 0.873 \\ \hline
Hybrid Grey Wolf - Whale Optimization Algorithm & \textbf{0.008} \\ \hline
\end{tabular}
\end{adjustbox}
\label{tab:ks_test_p_values}
\end{table}

\begin{figure*}[ht]
    \centering
    \includegraphics[width=0.8\textwidth]{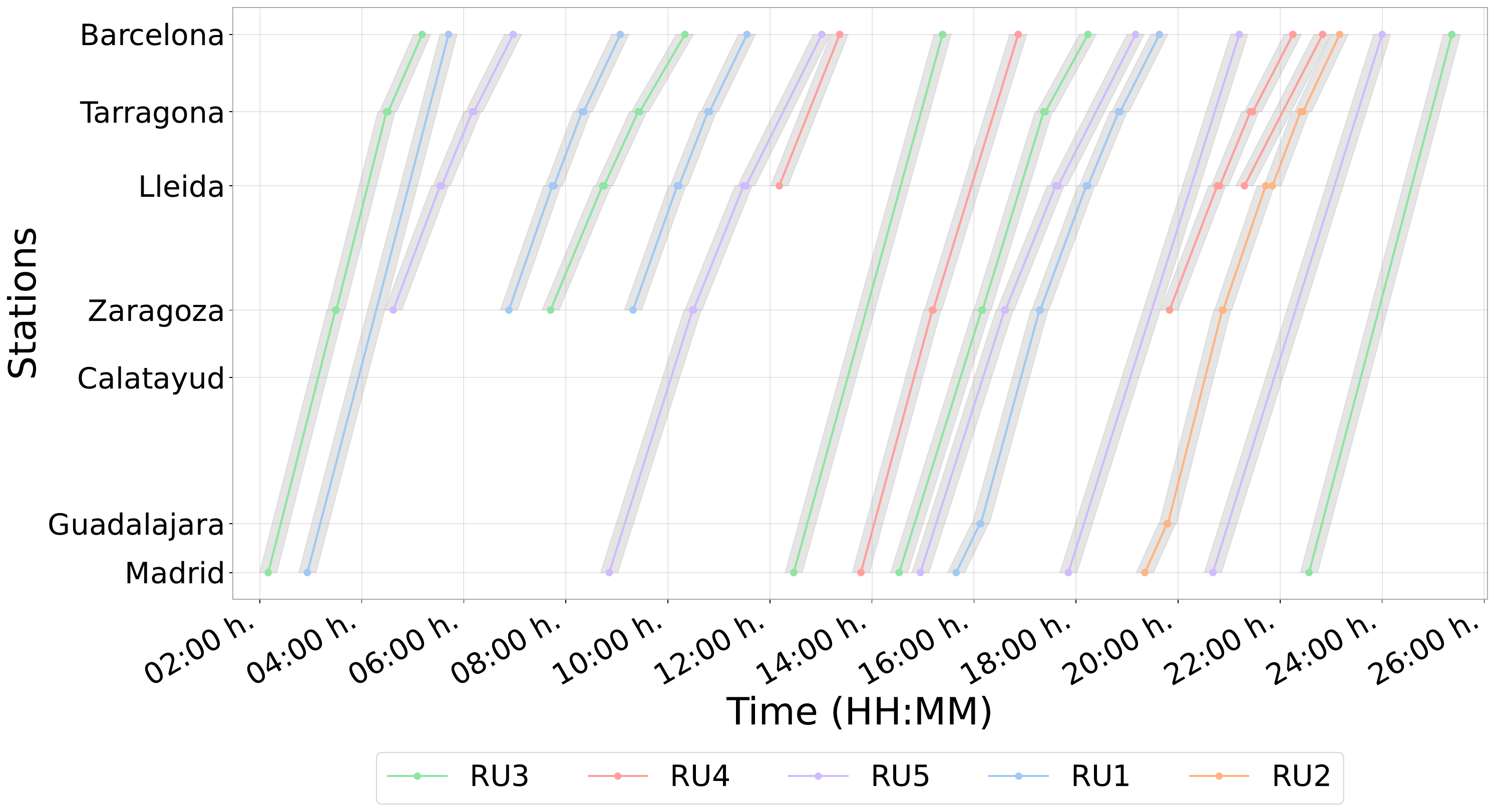}
    \caption{Marey chart with the solution returned by the \gls{GA} with hyperparameter tuning in the 4th run.}
    \label{fig:ga_marey_2} 
\end{figure*}

The results summarized in Table \ref{tab:ks_test_p_values} compare the performance of the 10 algorithms under two different conditions: runs without hyperparameter tuning versus runs with hyperparameter tuning. The Kolmogorov-Smirnov test for two-sample distributions was used to evaluate the statistical significance of the differences, taking a significance level of 95\%. The test was applied to the best fitness values obtained across five runs for each algorithm under both scenarios.The results show that for most algorithms, the differences in performance between the two scenarios are statistically significant. \gls{GA}, \gls{DE}, \gls{ACOR}, \gls{ABC}, \gls{GWO}, and \gls{GWO-WOA} all report p-values below the significance threshold, indicating significant improvements with hyperparameter tuning. These results underscore the sensitivity of these algorithms to parameter adjustments and the potential benefits of tuning. In contrast, \gls{PSO}, \gls{WOA}, and \gls{CMA-ES} exhibit much higher $p$-values, suggesting no significant differences between the two scenarios. For these algorithms, the default parameter configurations appear to perform similarly to the tuned settings, indicating robustness to hyperparameter changes or less sensitivity to the specific tuning applied. For this second experiment, the best solution is achieved with the \gls{GA} in the fourth run. The schedule obtained with this solution can be observed in Figure \ref{fig:ga_marey_2}. As with the first test, it can be observed how the algorithm makes the minimum adjustments required to meet safety constraints, while scheduling more services than originally requested. Again, some services could not be scheduled based on the restrictions established in the problem.

\subsection{Problem scalability}
\label{sec:problem_scalability}

This subsection assesses the scalability of the timetabling problem when addressed by two distinct approaches: metaheuristic algorithms using the \textit{MEALPY} library and a mathematical programming model implemented in Pyomo and solved with \gls{SCIP}. As the number of services and stations increases, both the combinatorial complexity of conflict detection and the size of the optimization model grow nonlinearly. Metaheuristic techniques rely on stochastic search mechanisms that can explore and exploit large solution spaces. In contrast, SCIP— currently one of the fastest solvers for Mixed Integer Programming (MIP) and Mixed Integer Nonlinear Programming (MINLP)— employs systematic branching, cutting planes, and presolving procedures to obtain optimal or near‐optimal schedules. By comparing execution times, convergence behavior, and solution quality across problem instances of increasing size, it becomes possible to determine how each methodology scales and to identify practical thresholds for real‐world applications.

A series of benchmark instances is created with 25 and 50 train services on a fixed network topology. For the metaheuristic approach, the Genetic Algorithm (GA) from the \textit{mealpy} library is set up with an optimized epoch count of 500 and a population size of 70. In contrast, the Pyomo SCIP formulation is solved with a wall‐clock time limit of 2 hours for the 25‐service instance and 8 hours for the 50‐service instance. For each instance, the following metrics are recorded: (a) total wall‐clock time to reach a feasible solution, (b) best revenue achieved within the time limit, and (c) optimality gap relative to SCIP’s best bound when available. This setup permits a direct comparison of GA’s stochastic search performance against SCIP’s exact optimization process as instance size grows.

The metaheuristic results for the 50‐service instance (Table \ref{tab:summary_results_50}) confirm that the Genetic Algorithm (GA) outperforms all other methods by a substantial margin. GA achieves the highest average revenue (\(\mu_{\text{revenue}} = 6721.32\)) and schedules nearly all services on average (\(\mu_{\text{trains}} = 29\) out of 50), with low variability (\(\sigma_{\text{revenue}} = 145.21\), \(\sigma_{\text{trains}} = 1\)). It finishes the optimization process (with 500 epochs and 70 individuals) in approximately 571 seconds on average. Differential Evolution (DE) is the second‐best, obtaining \(\mu_{\text{revenue}} = 5\,215.24\) and \(\mu_{\text{trains}} = 25\), though with higher spread (\(\sigma_{\text{revenue}} = 303.03\), \(\sigma_{\text{trains}} = 2\)) in about 483 seconds. Particle Swarm Optimization (PSO) ranks third (\(\mu_{\text{revenue}} = 4\,921.95\), \(\mu_{\text{trains}} = 19\)) but requires a longer runtime (\(\mu_{\text{time}} = 643\,\text{s}\)) and shows greater variability (\(\sigma_{\text{revenue}} = 460.84\), \(\sigma_{\text{trains}} = 2\)).

Overall, GA not only finds the highest‐quality schedules but does so with relatively low variance, making it the clear choice among metaheuristics for large‐scale timetable adjustment.  

\begin{table*}[htbp]
\centering
\caption{Summary of results with 50 requests.}
\begin{adjustbox}{max width=\textwidth}
\begin{tabular}{|c|l|c|c|c|c|c|c|}
\hline
\textbf{Rank} & \textbf{Algorithm} & \(\mu_{time}\) & \(\sigma_{time}\) & \(\mu_{revenue}\) & \(\sigma_{revenue}\) & \(\mu_{trains}\) & \(\sigma_{trains}\) \\ \hline \hline
1 & Genetic Algorithm                                     & 571.01 &   2.19 & \textbf{6721.32} & 145.21 & \textbf{29} & 1 \\ \hline
2 & Differential Evolution                                 & 483.30 &   1.69 & 5215.24 & 303.03 & 25 & 2 \\ \hline
3 & Particle Swarm Optimization                            & 643.06 &   4.78 & 4921.95 & 460.84 & 19 & 2 \\ \hline
4 & Ant Colony Optimization Continuous (ACOR)              & 220.95 &   0.87 & 4555.20 & 603.54 & 23 & 3 \\ \hline
5 & Covariance Matrix Adaptation Evolution Strategy        & 961.45 &  35.75 & 3997.22 & 168.33 & 19 & 1 \\ \hline
6 & Whale Optimization Algorithm                           & 505.12 &   4.71 & 3890.43 & 267.76 & 18 & 2 \\ \hline
7 & Grey Wolf Optimizer                                    & 776.15 &   1.81 & 3794.26 & 755.06 & 20 & 3 \\ \hline
8 & Artificial Bee Colony                                  & 863.28 &   1.09 & 3444.28 &  98.97 & 20 & 1 \\ \hline
9 & Hybrid Grey Wolf - Whale Optimization Algorithm        & 349.18 &   0.68 & 2735.24 &  \textbf{70.78} & 19 & 2 \\ \hline
10 & Simulated Annealing                                   &  73.06 &   0.98 & 2393.08 & 260.55 & 16 & 2 \\ \hline
\end{tabular}
\end{adjustbox}
\label{tab:summary_results_50}
\end{table*}

The Pyomo formulation includes nonlinear terms in the objective (penalizing deviations in departure and travel times via Equation \ref{ecu:penalty_curve} as well as bilinear products in the conflict‐avoidance constraints such as Equation \ref{eq:conflicts}. This combination of nonlinear penalties and bilinear scheduling variables makes the resulting model a mixed‐integer nonlinear program (MINLP) rather than a pure MIP.

SCIP was chosen because it provides native support for MINLPs, including nonlinear objectives and bilinear constraints, while still remaining non‐commercial. Its branch‐and‐bound engine, advanced presolving techniques, and integrated nonlinear solvers allow it to handle both the smooth penalty functions and the product terms in the conflict constraints. In practice, SCIP was able to produce high‐quality feasible solutions and tighten optimality gaps within the allotted time limits (2 hour for 25 services, 8 hour for 50 services), making it suitable for benchmarking against the GA’s stochastic search.

Table~\ref{tab:scip_25_summary} summarizes SCIP’s performance on the 25‐service instance. The original model comprises 83 variables (25 binary, 58 continuous) and 6973 constraints; after presolve, it expands to 383 variables (25 binary, 300 implicits, 58 continuous) and 6565 constraints (including 300 logical ``and'' and 6265 nonlinear rows). SCIP’s ALNS heuristic finds a nonzero solution ($ revenue = 445.48 $) within 1 second, and subsequent heuristics raise revenue to 4159.14 by 144 seconds, with over 67000 LP iterations. No finite dual bound is obtained before the 600 seconds limit (dual bound reported as \(+1\times10^{20}\)), so the optimality gap remains infinite. In practice, the large MINLP structure—dominated by nonlinear conflict and penalty constraints—prevents bound tightening, meaning SCIP produces high‐quality feasible schedules but cannot certify optimality within the time limit.

\begin{table}[htbp]
\centering
\scriptsize
\caption{SCIP performance summary on the 25‐service example.}
\begin{tabular}{|p{0.45\columnwidth}|c|}
\hline
\textbf{Metric} & \textbf{Value} \\ \hline
Original variables & 83 (25 binary, 58 continuous) \\ 
Original constraints & 6973 \\ \hline
Presolved variables & \makecell{383 (25 binary, \\ 300 implicits, 58 continuous)} \\ 
Presolved constraints & 6565 \\ 
  Nonlinear constraints & 6265 \\ 
  Logical “and” constraints & 300 \\ \hline
Best primal bound (revenue) & 4159.14 \\ 
Dual bound & \(+1\times10^{20}\) (no finite bound) \\ 
Optimality gap & \(\infty\) \\ \hline
Total LP iterations & 67669 \\ 
Branch-and-bound nodes processed & 8 \\ 
SCIP status & Time limit reached \\ 
Wall-clock solve time & 7200 (time limit) \\ \hline
\end{tabular}
\label{tab:scip_25_summary}
\end{table}

For the 50‐service case (Table~\ref{tab:scip_50_summary}), the model grows to 173 variables (50 binary, 123 continuous) and 26173 constraints, which presolving reduces to 1326 variables (50 binary, 1153 implicits, 123 continuous) and 25024 constraints (including 23 871 nonlinear and 1153 logical ``and'' rows). SCIP’s heuristics produce a best feasible revenue of 4145.98 but, even after 22625 branch‐and‐bound nodes and over 180 million LP iterations, no finite dual bound is found within the 36000-second limit (dual bound = \(+1\times10^{20}\), gap = $\infty$). The dramatic increase in nonlinear constraints and conflict terms further hampers bound tightening compared to the 25‐service case. Consequently, SCIP can still deliver comparable‐quality feasible schedules, but certifying optimality becomes even less practical as problem size doubles.

\begin{table}[htbp]
\centering
\scriptsize
\caption{SCIP performance summary on the 50‐service instance.}
\begin{tabular}{|p{0.45\columnwidth}|c|}
\hline
\textbf{Metric}                              & \textbf{Value}                                         \\ \hline
Original variables                           & 173 (50 binary, 123 continuous)                               \\ 
Original constraints                         & 26173                                                       \\ \hline
Presolved variables                          & \makecell{1326 (50 binary, 1\,153 implicits,\\123 continuous)} \\ 
Presolved constraints                        & 25024                                                        \\ 
\quad Nonlinear constraints                   & 23871                                                        \\ 
\quad Logical “and” constraints               & 1153                                                         \\ \hline
Best primal bound (revenue)                  & 4145.98                                                      \\ 
Dual bound                                    & \(+1\times10^{20}\) (no finite bound)                          \\ 
Optimality gap                               & \(\infty\)                                                     \\ \hline
Total LP iterations                          & 180172000                                                       \\ 
Branch‐and‐bound nodes processed              & 22625                                                         \\ 
SCIP status                                  & Time limit reached                                              \\ 
Wall‐clock solve time                        & 36000 (time limit)                                          \\ \hline
\end{tabular}
\label{tab:scip_50_summary}
\end{table}

The results in Table~\ref{tab:scalability} highlight clear trade‐offs between GA and Pyomo SCIP. In the 25‐service instance, SCIP proves marginally better revenue (4471.52 vs.\ 4457.33) but requires over 3000 seconds to reach that solution, whereas GA attains its best in only 174 seconds. Both methods schedule all 19 feasible trains, but GA's runtime is an order of magnitude lower—showing that metaheuristics can rapidly find near‐optimal schedules.

For the 50‐service instance, GA not only schedules a higher number trains (29) but also achieves significantly higher revenue (6838.61 vs.\ 4145.98). SCIP, limited by a 36000 seconds time cap, only schedules 21 trains and yields a much lower revenue. This contrast underscores that as problem size increases, SCIP's inability to close the optimality gap (and its slower convergence) results in suboptimal feasible solutions, whereas GA delivers higher revenue and better coverage in under 600 seconds.

\begin{table}[htbp]
\centering
\scriptsize
\caption{Scalability comparison of GA vs.\ Pyomo SCIP on 25 and 50 service requests.}
\begin{tabular}{|l|cc|cc|}
\hline
\multirow{2}{*}{\textbf{Metric}} & \multicolumn{2}{c|}{\textbf{25\,services}} & \multicolumn{2}{c|}{\textbf{50\,services}} \\ \cline{2-5}
& \textbf{GA}       & \textbf{SCIP}       & \textbf{GA}       & \textbf{SCIP}        \\ \hline
\multirow{2}{*}{\textbf{Variables}} & \multicolumn{2}{c|}{\textbf{Binary}: 25} & \multicolumn{2}{c|}{\textbf{Binary}: 50} \\ 
& \multicolumn{2}{c|}{\textbf{Continuous}: 58} & \multicolumn{2}{c|}{\textbf{Continuous}: 123} \\ \hline
\textbf{Best Revenue} & 4457.33 & \textbf{4471.52} & \textbf{6838.61} & 4145.98 \\ \hline
\textbf{Time to Best}  & \textbf{174.37s} & 53min & \textbf{570.11s} & 503min  \\ \hline
\textbf{Trains Scheduled} & 19 & 19 & \textbf{29} & 21 \\ \hline
\end{tabular}
\label{tab:scalability}
\end{table}

Overall, GA shows both computational efficiency and solution quality advantages at larger scales, while SCIP’s exact guarantees are feasible only for smaller instances and come at a substantial time cost.

\section{Conclusions}
\label{sec:Conclusions}

This study has explored the application of metaheuristic algorithms to the train timetabling problem in liberalized railway markets. The primary contributions of this work include the formulation of a mathematical model for timetabling optimization, a modular simulation framework, the evaluation of ten metaheuristic algorithms, and a computational analysis to determine their performance in a deregulated railway environment. The results show that \gls{GA} consistently outperformed other metaheuristics in terms of revenue optimization, convergence behavior, and adherence to requested schedules. Other algorithms such as \gls{PSO}, \gls{DE} and \gls{ACOR} also showed promising results, but exhibited higher variability and slower convergence. The study further revealed that the performance of metaheuristics can vary significantly depending on their configuration and the nature of the optimization problem, highlighting the importance of careful hyperparameter tuning. Specifically, statistical differences were observed between the results obtained with the tuned and non-tuned version of the following algorithms: \gls{GA}, \gls{DE}, \gls{ACOR}, \gls{ABC}, \gls{GWO} and \gls{GWO-WOA}. This improvement is mainly notable in the case of \gls{GA}, \gls{ACOR} and \gls{DE} as can be observed in the comparison between Table \ref{tab:summary_results_compact} and \ref{tab:summary_results_compactHIPER}. Despite achieving high revenues, the findings also underscored the trade-off between scheduling a higher number of trains and minimizing penalties due to schedule deviations. Algorithms such as \gls{GA} excelled at balancing these aspects, leading to very good results. The modular simulation framework developed for this study proved instrumental in evaluating different algorithms, and can be adapted for other optimization problems in railway systems.

Future research could expand on this work by exploring hybrid approaches that combine the strengths of multiple metaheuristics, thereby potentially improving both convergence speeds and solution quality. Additionally, integrating real-world data into the simulation framework could enhance the practical applicability of the proposed methods. Furthermore, an interesting line of research would be to consider the application of other nature-inspired computing and optimization algorithms. Specifically, the Harmony Search Algorithm, Water Drop Algorithms, Bacteria Foraging Algorithm, and Spider Monkey Optimization are highlighted. These alternative approaches can offer robust and efficient solutions aimed at improving convergence in complex optimization problems \cite{Siddique15}, adapting to dynamic optimization scenarios \cite{Siddique14a}, navigating complex search spaces \cite{Wang18}, addressing large-scale problems \cite{Akhand20}, enhancing search process efficiency and providing more robust solutions \cite{Siddique14b}, or improving efficiency in engineering applications \cite{Siddique16}. One further limitation of our study is its focus on a single, unidirectional high-speed corridor; as a result, the model neither distinguishes between main-line and branch-line track types nor explicitly handles shared-track segments. Extending the work to cover mixed-traffic operations and branch-line topologies would demand a comprehensive reformulation of both the objective function and the constraint set, with a richer problem representation to capture multiple line branches, bidirectional flows, and the nuances of shared infrastructure.


\section*{Acknowledgments}
This work was supported by grant PID2020-112967GB-C32 funded by \url{MCIN/AEI/10.13039/501100011033} and by ERDF A Way of Making Europe. It was completed when Enrique Adrian Villarrubia-Martin was a predoctoral fellow at Universidad de Castilla-La Mancha funded by the European Social Fund Plus (ESF+).

\bibliography{references}

\end{document}